\documentclass{article} 
\usepackage{iclr2024_conference,times}

\usepackage{amsmath,amsfonts,bm}

\def\figref#1{figure~\ref{#1}}

\def\secref#1{section~\ref{#1}}

\def\eqref#1{equation~\ref{#1}}

\def\1{\bm{1}}

\DeclareMathAlphabet{\mathsfit}{\encodingdefault}{\sfdefault}{m}{sl}
\SetMathAlphabet{\mathsfit}{bold}{\encodingdefault}{\sfdefault}{bx}{n}

\usepackage[hidelinks]{hyperref}
\usepackage{url}
\usepackage{bm}
\usepackage{amsmath}
\usepackage{amssymb}
\usepackage{multirow}
\usepackage{subfigure}
\usepackage{array}
\usepackage{amsthm}
\usepackage{rotating}
\usepackage{pifont}
\usepackage{wrapfig}
\usepackage{caption} 
\usepackage{tablefootnote}
\usepackage{savefnmark}
\usepackage{rotating}
\usepackage{tikz}
\usepackage{dcolumn, array, booktabs}
\usetikzlibrary{tikzmark, calc}
\usepackage{xcolor}

\renewcommand{\figref}[1]{Figure \ref{#1}}
\newcommand{\tabref}[1]{Table \ref{#1}}
\renewcommand{\secref}[1]{Section \ref{#1}}
\newcommand{\equref}[1]{Equation (\ref{#1})}

\let\oldcite\cite
\let\cite\citep
\let\citep\oldcite

\title{Extend Model Merging from Fine-Tuned to Pre-Trained Large Language Models via Weight Disentanglement}

\author{Le Yu, Bowen Yu, 
Haiyang Yu, Fei Huang, Yongbin Li\thanks{Corresponding author.}\\
  Alibaba Group \\
  \texttt{\{chuanyi.yl,yubowen.ybw,yifei.yhy,f.huang,shuide.lyb\}@alibaba-inc.com} 
}

\iclrfinalcopy 
\begin{document}

\maketitle

\begin{abstract}
Merging Large Language Models (LLMs) aims to amalgamate multiple homologous LLMs into one with all the capabilities. Ideally, any LLMs sharing the same backbone should be mergeable, irrespective of whether they are Fine-Tuned (FT) with minor parameter changes or Pre-Trained (PT) with substantial parameter shifts. However, existing methods often manually assign the model importance, rendering them feasible only for LLMs with similar parameter alterations, such as multiple FT LLMs. The diverse parameter changed ranges between FT and PT LLMs pose challenges for current solutions in empirically determining the optimal combination. In this paper, we make a pioneering effort to broaden the applicability of merging techniques from FT to PT LLMs. We initially examine the efficacy of current methods in merging FT and PT LLMs, discovering that they struggle to deal with PT LLMs. Subsequently, we introduce an approach based on \textbf{W}e\textbf{I}ght \textbf{D}is\textbf{EN}tanglement (WIDEN) to effectively extend the merging scope, which first disentangles model weights into magnitude and direction components, and then performs adaptive fusion by considering their respective contributions. In the experiments, we merge Qwen1.5-Chat (an FT LLM with instruction-following skills) with Sailor (a PT LLM with multilingual abilities) across 7B and 14B model scales. Results reveal that: (1) existing solutions usually fail when merging Sailor, either losing both abilities or only retaining instruction-following skills; (2) WIDEN successfully injects the multilingual abilities of Sailor into Qwen1.5-Chat and make it proficient in Southeast Asian languages, achieving enhancements in the fundamental capabilities. In light of previous research, we also merge multiple 13B FT LLMs and observe that WIDEN achieves a balanced amalgamation of instruction following, mathematical reasoning, and code generation skills.
\end{abstract}

\section{Introduction}
\label{section-1}

In recent years, model merging has sparked significant interest as a prominent topic, which intends to integrate multiple homologous models (sharing the same backbone) into a singular one that encapsulates all the abilities \cite{DBLP:conf/icml/WortsmanIGRLMNF22,DBLP:conf/nips/MatenaR22,DBLP:conf/iclr/IlharcoRWSHF23,DBLP:conf/iclr/Jin0P023,DBLP:journals/corr/abs-2403-19522,DBLP:conf/nips/YadavTCRB23,DBLP:journals/corr/abs-2312-06795,yu2023language}. Distinct from other approaches that can also amalgamate various skills (e.g., ensemble learning \cite{mohammed2023comprehensive}, multi-task learning \cite{DBLP:journals/corr/abs-2009-09796,DBLP:journals/tkde/ZhangY22}), model merging is lauded for its computational frugality, especially when applied to Large Language Models (LLMs). Notably, it achieves integration without using additional training data or even GPUs, establishing a new paradigm for efficiently combining LLMs' capabilities \cite{yu2023language}.

\begin{figure}[!htbp]
    \centering
    \begin{minipage}{0.56\textwidth}
        \centering
        \includegraphics[width=\columnwidth]{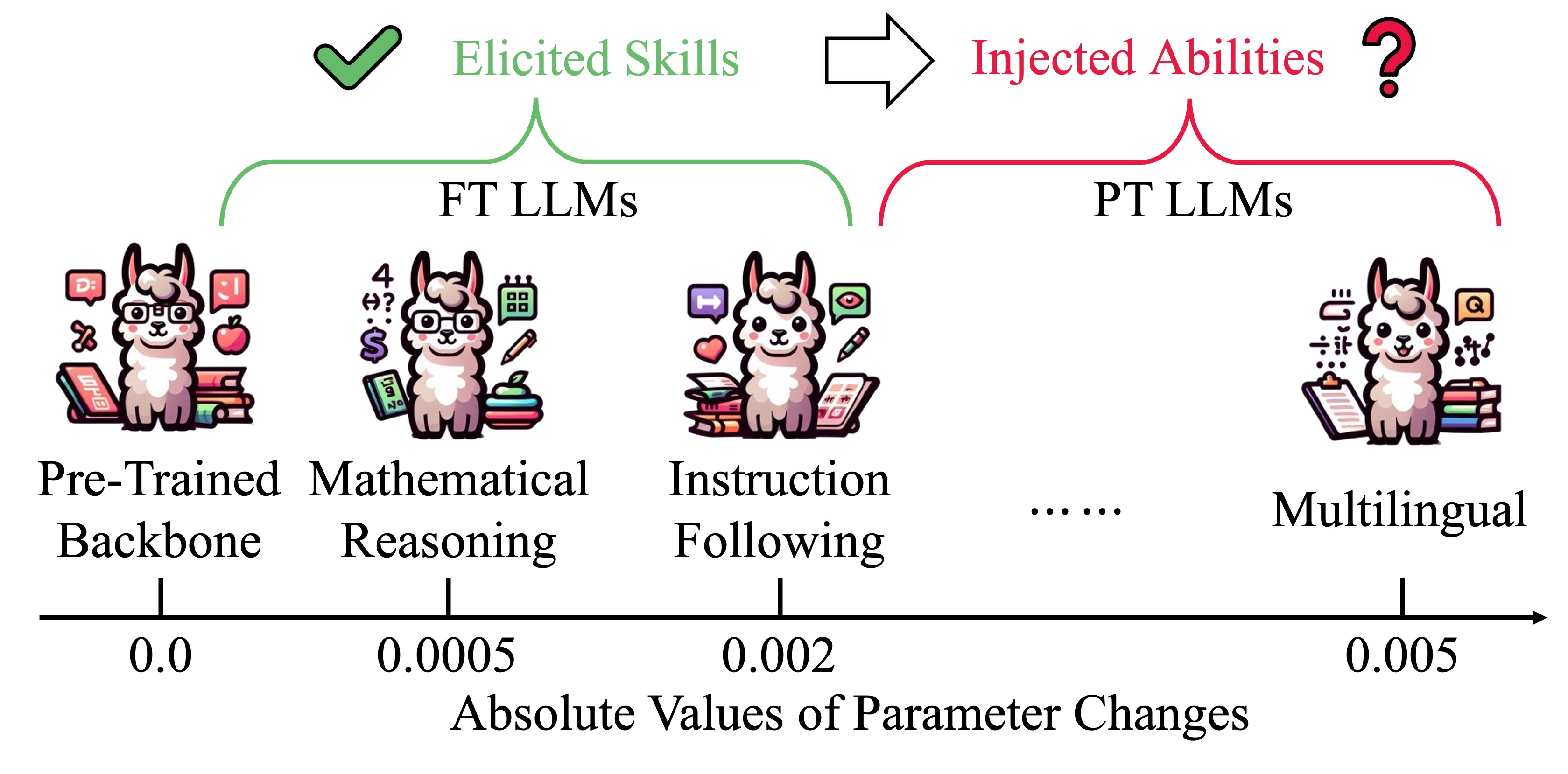}
        \caption{Issues of existing merging techniques.}
        \label{fig:motivation}
    \end{minipage}\hfill
    \begin{minipage}{0.44\textwidth}
        \centering
        \captionof{table}{Average results of merging Qwen1.5-14B-Chat and Sailor-14B. Metrics of the best methods in Arithmetic, Geometric, and Pruning categories are reported.}
        \label{tab:14B_llm_merge_comparison}
        \resizebox{\textwidth}{!}
        {
        \setlength{\tabcolsep}{0.8mm}
        {
        \begin{tabular}{c|cc}
        \hline
        & \begin{tabular}[c]{@{}c@{}}Instruction \\ Following\end{tabular} & Multilingual   \\ \hline
        Qwen1.5-14B-Chat & 68.08                                                            & 53.74          \\
        Sailor-14B       & 64.02                                                            & 59.90          \\ \hline
        Arithmetic-based & 66.30 (-1.78)                                                    & 40.72 (\textcolor{red}{-19.18}) \\
        Geometric-based  & 67.59 (-0.49)                                                    & 49.52 (\textcolor{red}{-10.38}) \\
        Pruning-based    & 51.72 (\textcolor{red}{-16.36})                                                   & 28.69 (\textcolor{red}{-31.21}) \\
        WIDEN            & 66.75 (-1.33)                                                    & 59.67 (-0.23)  \\ \hline
        \end{tabular}
        }
        }
    \end{minipage}
\end{figure}

Technically, there are predominantly two strategies to equip LLMs with desired capabilities \cite{DBLP:journals/corr/abs-2303-18223}: fine-tuning to elicit existing skills \cite{DBLP:journals/corr/abs-2307-12966,DBLP:journals/corr/abs-2308-10792} and pre-training to inject new abilities \cite{DBLP:journals/corr/abs-2402-01364}. Existing merging methods mainly focus on integrating the skills of Fine-Tuned (FT) LLMs with minor parameter changes relative to the backbone, typically within 0.002 \cite{yu2023language}. However, it is crucial to acknowledge that pre-training is the cornerstone for fundamentally enhancing the capabilities of LLMs. The practicality of merging techniques in scenarios where Pre-Trained (PT) LLMs undergo substantial parameter shifts remains unexplored, as depicted in \figref{fig:motivation}. Consequently, if the application of merging is restricted to FT LLMs, its potential for broader improvement would be significantly constrained.

To fill in the aforementioned blank, this work makes two key technical contributions. 

\textbf{We examine the feasibility of existing approaches in absorbing the abilities from PT LLMs}. We investigate the performance of widely used arithmetic-based \cite{DBLP:conf/icml/WortsmanIGRLMNF22,DBLP:conf/iclr/IlharcoRWSHF23}, geometric-based \cite{DBLP:conf/siggraph/Shoemake85,DBLP:journals/corr/abs-2403-19522}, and pruning-based \cite{DBLP:conf/nips/YadavTCRB23,DBLP:journals/corr/abs-2312-06795,yu2023language} methods when merging FT and PT LLMs. As illustrated in \tabref{tab:14B_llm_merge_comparison}, we find current methods either lose efficacy in retaining the abilities of PT LLMs (leading to a decrease of approximately 10 to 20 points on average) or fail to preserve both capabilities (resulting in an average degradation of about 15 and 30 points, respectively). One possible reason is that existing methods depend on manually assigned scaling terms to gauge the model contribution, which is only applicable when multiple LLMs depict comparable parameter alterations. Nonetheless, when confronted with diverse parameter changed ranges between FT and PT LLMs, deriving the optimal scaling factors according to human expertise becomes exceedingly arduous.

\textbf{We propose a new solution grounded in WeIght DisENtanglement (WIDEN) to expand the scope of merging techniques from FT to PT LLMs}. WIDEN tackles the drawbacks of existing works by automatically computing the model importance in the merging process without requiring manual specification, mitigating the influence induced by diverse parameter changed ranges between FT and PT LLMs. To be specific, WIDEN first disentangles each weight of a given LLM into two components: \textit{magnitude} and \textit{direction}. Then, the divergence of each component relative to the backbone is quantified to provide a numerical measure of how much each LLM has been altered. Next, WIDEN employs a ranking mechanism within each LLM to obtain the weight importance, tackling the diversity in parameter changed ranges between FT and PT LLMs. Finally, WIDEN performs adaptive merging on multiple LLMs by Softmax with the score calibration design.

We experiment with Qwen1.5-Chat \cite{DBLP:journals/corr/abs-2309-16609} (an FT LLM with instruction-following skills) and Sailor \cite{DBLP:journals/corr/abs-2404-03608} (a PT LLM with multilingual abilities for South-East Asia) across 7B and 14B model scales to verify the effectiveness of WIDEN for model merging\footnote{To the best of our knowledge, Sailor is one of the few publicly accessible LLM that has undergone extensive continued pre-training upon the open-source Qwen1.5 model, ideally suitable to our experimental scenarios. Therefore, Sailor and its homologous counterpart, Qwen1.5-Chat, are selected for our study.}. Experimental results indicate that WIDEN outperforms existing methods by not only absorbing the multilingual abilities of Sailor but also preserving the instruction-following skills of Qwen1.5-Chat. For example, in \tabref{tab:14B_llm_merge_comparison}, WIDEN slightly causes an average reduction of 0.23 and 1.33 points for Sailor-14B and Qwen1.5-14B-Chat, respectively. These observations demonstrate that WIDEN effectively extends the applicability of merging techniques from FT to PT LLMs. Considering previous works, we further merge three FT LLMs including WizardLM-13B \cite{xu2024wizardlm} for instruction following, WizardMath-13B \cite{DBLP:journals/corr/abs-2308-09583} for mathematical reasoning, and llama-2-13b-code-alpaca \cite{codealpaca} for code generation. Results show that WIDEN is also feasible under the conventional setting and can strike a favorable balance among these capabilities.

Resources are available at \url{https://github.com/yule-BUAA/MergeLLM}.

\section{Related Work}
\label{section-2}

\textbf{Fine-Tuning and Pre-Training of LLMs}. 
Generally, LLMs can be adapted to various tasks via two strategies: fine-tuning and pre-training \cite{DBLP:journals/corr/abs-2303-18223}. Fine-tuning is designed to elicit backbones with specific skills by optimizing them on a limited set of task-specific data, obtaining FT LLMs with skills such as instruction following \cite{DBLP:conf/nips/RafailovSMMEF23,DBLP:conf/aaai/00010LYHLW24} and mathematical reasoning \cite{DBLP:journals/corr/abs-2308-01825,DBLP:journals/corr/abs-2308-09583}. The fine-tuning process typically brings minor modifications to the model parameters \cite{yu2023language}, holding true for both full fine-tuning approaches \cite{radford2018improving,DBLP:conf/naacl/DevlinCLT19} and parameter-efficient fine-tuning techniques \cite{DBLP:conf/icml/HoulsbyGJMLGAG19,DBLP:conf/acl/LiL20,DBLP:conf/emnlp/LesterAC21,DBLP:conf/iclr/HuSWALWWC22}.
In contrast to fine-tuning, pre-training trains LLMs on large-scale raw corpora to enhance models with domain knowledge \cite{DBLP:conf/emnlp/KeLS0SL22,DBLP:conf/iclr/KeSLKK023,DBLP:journals/corr/abs-2309-09530}, deriving PT LLMs with fundamental abilities like finance analysis \cite{DBLP:journals/corr/abs-2311-08545} and law assistance \cite{colombo2024saullm}. Pre-training often leads to more obvious parameter shifts than fine-tuning due to extensive data used during the phase. Different from current merging methods that are only applicable to FT LLMs, this paper proposes a new solution to innovatively harness the capabilities of PT LLMs.

\textbf{Merging of LLMs}.
Model merging aims to amalgamate multiple homologous models (derived from the same backbone) into a single one that possesses all the abilities \cite{DBLP:conf/icml/WortsmanIGRLMNF22,DBLP:conf/nips/MatenaR22,DBLP:conf/iclr/IlharcoRWSHF23,DBLP:conf/iclr/Jin0P023,DBLP:journals/corr/abs-2403-19522,DBLP:conf/nips/YadavTCRB23,DBLP:journals/corr/abs-2312-06795,yu2023language}. The allure of the model merging technique stems from its minimal computational expense, particularly favorable for LLMs, which can be realized without retraining or GPUs \cite{yu2023language}. 
Existing merging techniques that are feasible for LLMs can be broadly categorized into three groups, which are based on arithmetic, geometric, and pruning. Average Merging \cite{DBLP:conf/icml/WortsmanIGRLMNF22} and Task Arithmetic \cite{DBLP:conf/iclr/IlharcoRWSHF23} belong to arithmetic-based approaches. The former utilizes averaged parameters to create the merged model, whereas the latter introduces the concept of task vector (i.e., parameter difference between an FT model and its backbone) and uses a scaling term to regulate the importance of various models. As geometric-based methods, both SLERP \cite{DBLP:conf/siggraph/Shoemake85} and Model Stock \cite{DBLP:journals/corr/abs-2403-19522} consider the geometric properties in weight space. In particular, SLERP is specifically designed for the integration of two models, which performs spherical interpolation of model weights. Model Stock approximates a center-close weight based on several FT models, utilizing their backbone as an anchor point. TIES-Merging \cite{DBLP:conf/nips/YadavTCRB23}, Breadcrumbs \cite{DBLP:journals/corr/abs-2312-06795}, and DARE \cite{yu2023language} are methods based on pruning. TIES-Merging eliminates parameter interference among multiple models by first removing delta parameters with low magnitudes and then merging parameters with consistent signs after resolving disagreements. Breadcrumbs masks out the extreme tails (also known as outliners) of the absolute magnitude distribution of task vectors to obtain the final model. DARE is a versatile plug-in for existing merging approaches, which first randomly drops delta parameters and then rescales the remaining ones to maintain model performance. However, most of the current methods manually determine the importance of each model, suitable only for LLMs with similar parameter changes. When the parameter changed ranges are diverse between FT and PT LLMs, determining the optimal combination becomes overwhelmingly challenging. This paper initially verifies the limitations of existing methods in combining the abilities of PT LLMs. Subsequently, an approach based on weight disentanglement is introduced to effectively expand the scope of merging techniques from FT to PT LLMs.

\section{Methodology}
\label{section-3}

\subsection{Preliminaries}
\textbf{Merging Beyond FT LLMs}. Given a collection of $N$ homologous LLMs characterized by parameters $\left\{\Theta^1, \Theta^2, \cdots, \Theta^N\right\}$, all of which share the same backbone with parameters $\Theta_{\text{PRE}}$, model merging aims to amalgamate the parameters of $N$ LLMs into a singular model with all the capabilities, denoted as $\Theta_{\text{M}}$. Previous studies only focus on combining the skills of FT LLMs parameterized by $\left\{\Theta_{\text{FT}}^1, \Theta_{\text{FT}}^2, \cdots, \Theta_{\text{FT}}^N\right\}$, where each model exhibits slight parameter changes, usually within 0.002 \cite{yu2023language}. In this paper, we extend the scope of merging techniques from FT to PT LLMs, intending to absorb the abilities of PT LLMs. Therefore, the parameters targeted for merging become $\left\{\Theta_{\text{TYPE}_1}^1, \Theta_{\text{TYPE}_2}^2, \cdots, \Theta_{\text{TYPE}_N}^N\right\}$, where $\text{TYPE}_n$ ($1 \leq n \leq N$) can be either FT or PT.

\textbf{Weight Disentanglement}. As outlined in \citet{DBLP:conf/nips/SalimansK16,DBLP:journals/corr/abs-2402-09353}, a weight $\bm{W} \in \mathbb{R}^{d \times k}$ can be disentangled into two components: a row vector $\bm{m} \in \mathbb{R}^{1 \times k}$ that captures the magnitudes and a matrix $\bm{D} \in \mathbb{R}^{d \times k}$ that stores the direction vectors. Here, $d$ and $k$ represent the output and input dimensions. Mathematically, the disentanglement of weight $\bm{W}$ is achieved by
\begin{equation}
    \label{equ:weight_disentanglement}
    \bm{W} = \bm{m} \bm{D} = \|\bm{W}\|_c \frac{\bm{W}}{\|\bm{W}\|_c},
\end{equation}
where $\|\cdot\|_c$ denotes the vector-wise $l_c$-norm of a matrix across each column. Such a decoupling operation guarantees that each column $\bm{D}_{:,j}$ ($1\leq j \leq k$) is a unit vector, and scalar $m_j \in \bm{m}$ signifies the magnitude of direction vector $\bm{D}_{:,j}$. Since the primary challenge of extending merging scope to PT LLMs lies in the manual assignment of model importance, we employ weight disentanglement to initially decouple weights into magnitudes and directions, and then automatically compute the weight importance without human expertise based on these two components.

\subsection{Exploring Efficacy of Current Methods When Merging PT LLMs}\label{section-3-1-baseline_performance_for_merging_PT_LLMs}
We investigate the efficacy of seven commonly used merging techniques when integrating the abilities of PT LLMs. To be specific, Average Merging \cite{DBLP:conf/icml/WortsmanIGRLMNF22} and Task Arithmetic \cite{DBLP:conf/iclr/IlharcoRWSHF23} are arithmetic-based methods. SLERP \cite{DBLP:conf/siggraph/Shoemake85} and Model Stock \cite{DBLP:journals/corr/abs-2403-19522} belong to geometric-based approaches. TIES-Merging \cite{DBLP:conf/nips/YadavTCRB23}, Breadcrumbs \cite{DBLP:journals/corr/abs-2312-06795} and DARE \cite{yu2023language} are pruning-based solutions. Please see \secref{section-appendix-model_merging_methods_descriptions} for detailed descriptions of these methods. To evaluate the performance, we attempt to combine the instruction-following skills of an FT LLM, Qwen1.5-Chat \cite{DBLP:journals/corr/abs-2309-16609}, and the multilingual abilities of a PT LLM, Sailor \cite{DBLP:journals/corr/abs-2404-03608}. Experimental setup, results, and analysis can be found in \secref{section-4}. 

Since this part mainly concentrates on the feasibility of merging techniques when applied to PT LLMs, we highlight the key conclusion pertinent to PT LLMs: \textit{existing merging approaches face difficulties in preserving the abilities of PT LLMs}. As evidenced in \tabref{tab:llms_merging_southeast_asian_benchmark}, the performance of all merging methods on the multilingual abilities significantly declines. This phenomenon is largely attributed to the reliance of most methods on manually assigned scaling factors to determine the contribution of each model at various levels throughout the merging process, encompassing model level \cite{DBLP:conf/iclr/IlharcoRWSHF23,DBLP:conf/nips/YadavTCRB23,DBLP:journals/corr/abs-2312-06795}, layer/module level \cite{DBLP:journals/corr/abs-2403-13257}, and parameter level \cite{DBLP:conf/siggraph/Shoemake85}. The diverse parameter changed ranges between FT and PT LLMs complicate the manual assignment of model importance, making it intractable to define optimal scaling factors case by case.

\subsection{Extending Merging Scope to PT LLMs via Weight Disentanglement}
We present a new approach based on \textbf{W}e\textbf{I}ght \textbf{D}is\textbf{EN}tanglement (WIDEN) to innovatively broaden the applicability of model merging techniques from FT to PT LLMs, whose key concept is to adaptively assess the importance of weights during the merging process for neutralizing the effects of diverse parameter changed ranges between FT and PT LLMs. The framework of WIDEN comprises four main steps. Given the weights of LLMs (including the backbone as well as models to be merged), WIDEN 1) disentangles each weight into a row vector of magnitudes and a matrix of direction vectors; 2) estimates weight divergence relative to the backbone founded on absolute values of magnitude alterations and cosine similarities between direction vectors; 3) ranks the weights inside each LLM grounded in their divergence to derive the weight importance, thereby mitigating the impact of diverse parameter changed ranges; 4) merges multiple LLMs into a single one according to the obtained weight importance via Softmax with score calibration.

\textbf{Disentangling Weights of LLMs}. 
Given multiple homologous LLMs (each LLM can be obtained by either FT or PT) with parameters $\left\{\Theta^1, \Theta^2, \cdots, \Theta^N\right\}$ as well as the backbone with parameters $\Theta_{\text{PRE}}$, we first perform weight disentanglement for the parameters. Take $\bm{W}^n \in \Theta^n$ with shape $\mathbb{R}^{d \times k}$ as an example\footnote{Note that $\Theta^n$ represents the collection of parameters of the $n$-th LLM, consisting of a multitude of weights.}. $\bm{W}^n$ can be decoupled into $\bm{m}^n = \|\bm{W}^n\|_c \in \mathbb{R}^{1 \times k}$ and $\bm{D}^n = \frac{\bm{W}^n}{\|\bm{W}^n\|_c} \in \mathbb{R}^{d \times k}$. After applying this disentanglement across all the LLMs, we can obtain the sets of row vectors of magnitudes $\{\bm{m}^n\}_{n=1}^N \cup \{\bm{m}_{\text{PRE}}\}$ and matrices of direction vectors $\{\bm{D}^n\}_{n=1}^N \cup \{\bm{D}_{\text{PRE}}\}$.

\textbf{Estimating Weight Divergence Relative to Backbone}.
We estimate the weight divergence of each LLM relative to the backbone from the perspective of magnitudes and directions with two measurements. To be specific, we compute the absolute values of magnitude alterations and determine the changes between direction vectors based on cosine similarities as follows,
\begin{equation}
\begin{split}
   & \Delta \bm{m}^n = | \bm{m}^n\ - \bm{m}_{\text{PRE}} | \in \mathbb{R}^{1 \times k}, \ \text{for} \ 1 \leq n \leq N, \\
   \Delta D_j^n = 1 - & \text{CosineSimilarity}(\bm{D}_{:,j}^n, {\bm{D}_{\text{PRE}}}_{:,j}) \in \mathbb{R}, \ \text{for} \ 1 \leq j \leq k, \ 1 \leq n \leq N,
\end{split}
\end{equation}
where $\text{CosineSimilarity}(\bm{x}, \bm{y}) = \frac{\bm{x} \cdot \bm{y}}{\|\bm{x}\|_2 \cdot \|\bm{y}\|_2}$. Thus, we obtain the divergences of the LLMs relative to the backbone in both magnitudes $\{\Delta \bm{m}^n \in \mathbb{R}^{1 \times k}\}_{n=1}^N$ and directions $\{\Delta \bm{D}^n \in \mathbb{R}^{1 \times k}\}_{n=1}^N$.

\textbf{Ranking Weights Inside Each LLM}.
We design a ranking mechanism to alleviate the potential impact of diverse parameter changed ranges among various LLMs, which assigns importance to the weights within each LLM according to their divergence relative to the backbone (greater divergence indicates higher essentiality). The ranking mechanism is applied to both the magnitudes and the directions of weights. To illustrate, consider the magnitudes as an instance. Given $\Delta \bm{m}^n \in \mathbb{R}^{1 \times k}$ of the $n$-th LLM, we initially sort $\Delta \bm{m}^n$ in ascending order, yielding an index row vector $\bm{m}_{\text{IND}}^n \in \mathbb{R}^{1 \times k}$ that contains values ranging from 1 to $k$. Subsequently, we derive a row vector $\widetilde{\bm{m}}^n \in \mathbb{R}^{1 \times k}$ that encapsulates normalized ranking scores based on $\bm{m}_{\text{IND}}^n$, which is computed by
\begin{equation}
   {\widetilde{m}^n}_{{m_{\text{IND}}^n}_j} = j / k, \ \text{for} \ 1 \leq j \leq k.
\end{equation}
$\widetilde{\bm{m}}^n \in \mathbb{R}^{1 \times k}$ represents the normalized importance of each position within the range $[1, \cdots, k]$ for the $n$-th LLM. Following the same procedure, the directions of weights can also be assigned with normalized importance, which can be denoted by $\widetilde{\bm{D}}^n \in \mathbb{R}^{1 \times k}$. Such a ranking mechanism ensures that, within each LLM, the importance of magnitudes and directions is uniformly distributed between 0 and 1, thereby eliminating the potential influences arising from diverse parameter changed ranges between FT and PT LLMs. After applying the ranking operation for all the LLMs, we can ultimately obtain $\{\widetilde{\bm{m}}^n \in \mathbb{R}^{1 \times k}\}_{n=1}^N$ and $\{\widetilde{\bm{D}}^n \in \mathbb{R}^{1 \times k}\}_{n=1}^N$.

\textbf{Merging LLMs via Softmax with Score Calibration}.
We employ an adaptive merging strategy for multiple LLMs through a Softmax function, complemented by score calibration. Initially, we calculate the importance scores for magnitudes and directions by applying the Softmax function to $\{\widetilde{\bm{m}}^n \in \mathbb{R}^{1 \times k}\}_{n=1}^N$ and $\{\widetilde{\bm{D}}^n \in \mathbb{R}^{1 \times k}\}_{n=1}^N$, yielding $\widetilde{\bm{\mathcal{M}}}, \widetilde{\bm{\mathcal{D}}} \in \mathbb{R}^{N \times k}$ by
\begin{equation}
\begin{split}
    \widetilde{\mathcal{M}}_{n,j} = \frac{\exp{(\widetilde{m}^n_j)}}{\sum_{n^\prime=1}^N \exp{(\widetilde{m}^{n^\prime}_j)}} \in \mathbb{R}, \ 
    \widetilde{\mathcal{D}}_{n,j} = \frac{\exp{(\widetilde{D}^n_j)}}{\sum_{n^\prime=1}^N \exp{(\widetilde{D}^{n^\prime}_j)}} \in \mathbb{R}, \ \text{for} \ 1 \leq j \leq k, \ 1 \leq n \leq N,
\end{split}
\end{equation}
However, Softmax restricts the sum of parameter importance across multiple LLMs to 1, potentially diminishing the significance of crucial parameters in certain cases. Thus, we incorporate a score calibration operation to relax the constraint of Softmax for essential parameters. We identify crucial parameters as those whose importance exceeds the average level by a factor of $t$ as follows,
\begin{equation}
    \label{equ:weight_above_average}
   \mathbb{P}^n_m = \{j | \widetilde{m}^n_j > \frac{t}{k} \cdot \sum\nolimits_{j^\prime=1}^k \widetilde{m}^n_{j^\prime} \}, \ \mathbb{P}^n_D = \{j | \widetilde{D}^n_j > \frac{t}{k} \cdot \sum\nolimits_{j^\prime=1}^k \widetilde{D}^n_{j^\prime} \}.
\end{equation}
Subsequently, we calibrate the scores using $\mathbb{P}^n_m$ and $\mathbb{P}^n_D$ by
\begin{align}
\begin{aligned}
\mathcal{M}_{n,j} &=
\begin{cases} 
    s, & \text{if } j \in \mathbb{P}^n_m \\
    \widetilde{\mathcal{M}}_{n,j}, & \text{if } j \notin \mathbb{P}^n_m
\end{cases}
\end{aligned}
, \ \
\begin{aligned}
\mathcal{D}_{n,j} &=
\begin{cases} 
    s, & \text{if } j \in \mathbb{P}^n_D \\
    \widetilde{\mathcal{D}}_{n,j}, & \text{if } j \notin \mathbb{P}^n_D
\end{cases}
\end{aligned}
, \ \
\end{align}
where $s$ regulates the numerical value of score calibration. Finally, we integrate the weights of multiple LLMs into $\bm{W}_\text{M}$ by considering the adjusted contributions of both magnitudes and directions,
\begin{equation}
   \label{equ:weight_adaptive_merging}
   \bm{W}_\text{M} = \bm{W}_\text{PRE} + \sum\limits_{n=1}^N \frac{\bm{\mathcal{M}}_{n,:} + \bm{\mathcal{D}}_{n,:}}{2} \odot \left( \bm{W}^n - \bm{W}_\text{PRE} \right) \in \mathbb{R}^{d \times k}.
\end{equation}

\textit{Remark 1}. The aforementioned procedure is designed to deal with two-dimensional weights within LLMs, accounting for both magnitudes and directions. For one-dimensional parameters, such as weights in normalization layers and biases in linear transformations, we handle them as vectors of magnitudes and estimate their changes relative to the backbone by absolute values of the differences.

\textit{Remark 2}. Existing arithmetic-based merging methods including Average Merging \cite{DBLP:conf/icml/WortsmanIGRLMNF22} and Task Arithmetic \cite{DBLP:conf/iclr/IlharcoRWSHF23}, can be viewed as special instances of WIDEN. Specifically, when $t < 0.0$ and $s = 1/N$, WIDEN aligns with the principles of Average Merging. When $t < 0.0$ and $s = \lambda$ ($\lambda$ is the scaling term in Task Arithmetic), WIDEN becomes Task Arithmetic. Please refer to \secref{section-appendix-analysis_widen_arithmetic} for the detailed analysis.

\section{Experiments}
\label{section-4}
We conduct experiments on model merging in two scenarios: 1) integrating both FT and PT LLMs, a new setting not explored before; 2) combining FT LLMs as in previous research.

\subsection{Experimental Setup}\label{section-4-experimental_setup}
\textbf{Merging FT and PT LLMs}. 
We choose Qwen1.5-Chat \cite{DBLP:journals/corr/abs-2309-16609} with instruction-following skills as the FT LLM and select Sailor \cite{DBLP:journals/corr/abs-2404-03608} with multilingual abilities for South-East Asia as the PT LLM. Both models adopt Qwen1.5 \cite{DBLP:journals/corr/abs-2309-16609} as the backbone. Open LLM Leaderboard \cite{open-llm-leaderboard} and benchmark for South-East Asian languages \cite{DBLP:journals/corr/abs-2404-03608} are used for evaluating the performance of models across 7B and 14B sizes.

\textbf{Merging FT LLMs}. 
In accordance with \citet{yu2023language}, we merge three FT LLMs that are based on Llama-2-13b \cite{DBLP:journals/corr/abs-2307-09288}: WizardLM-13B \cite{xu2024wizardlm} for instruction following, WizardMath-13B \cite{DBLP:journals/corr/abs-2308-09583} for mathematical reasoning, and llama-2-13b-code-alpaca \cite{codealpaca} for code generation. AlpacaEval 2.0 \cite{DBLP:journals/corr/abs-2404-04475}, GSM8K \cite{DBLP:journals/corr/abs-2110-14168}, MATH \cite{DBLP:conf/nips/HendrycksBKABTS21}, HumanEval \cite{DBLP:journals/corr/abs-2107-03374}, and MBPP \cite{DBLP:journals/corr/abs-2108-07732} are utilized for evaluation. 

Please see \secref{section-appendix-datasets_metrics_details} for the overview and evaluation metrics of the benchmarks. Also, refer to \tabref{tab:llms_backbone_correspondences} in \secref{section-appendix-llms_backbone_correspondences} for the details of FT and PT LLMs. We compare WIDEN with seven popular baselines for model merging, including Average Merging \cite{DBLP:conf/icml/WortsmanIGRLMNF22}, Task Arithmetic \cite{DBLP:conf/iclr/IlharcoRWSHF23}, SLERP \cite{DBLP:conf/siggraph/Shoemake85}, Model Stock \cite{DBLP:journals/corr/abs-2403-19522}, TIES-Merging \cite{DBLP:conf/nips/YadavTCRB23}, Breadcrumbs \cite{DBLP:journals/corr/abs-2312-06795}, and DARE \cite{yu2023language}. See \secref{section-3-1-baseline_performance_for_merging_PT_LLMs} and \secref{section-appendix-model_merging_methods_descriptions} for more descriptions.

\textbf{Configurations of Merging Methods}.
We apply grid search to identify the optimal settings for various merging techniques. The proposed WIDEN utilizes $l_2$ normalization and involves two hyperparameters: $s$ and $t$. For ease of implementation, the score calibration factor $s$ is consistently fixed to 1.0 across all the cases. The factor $t$ is determined by grid search. Please refer to \tabref{tab:hyperparameter_searched_ranges_merging_methods} in \secref{section-appendix-grid_search_model_merging_hyperparameters} for detailed information about the searched ranges.

\textbf{Hardware Requirements}. The process of merging LLMs requires only CPU resources. To evaluate the merged LLMs, we employ A100 GPUs equipped with 80 GB of memory. Notably, all the experiments can be successfully reproduced using a single A100 GPU.

\subsection{Performance of Merging FT and PT LLMs}
\tabref{tab:llms_merging_southeast_asian_benchmark} shows the results of merging Qwen1.5-Chat and Sailor on South-East Asian language benchmark. Since Average Merging is a special case of Task Arithmetic when the scaling term is 0.5, we thereby only report the results of Task Arithmetic, which inherently include the performance of Average Merging. Note that th, id, vi, and jv are abbreviations of Thai, Indonesian, Vietnamese, and Javanese. The best and second-best results are marked in \textbf{bold} and \underline{underlined} fonts. From \tabref{tab:llms_merging_southeast_asian_benchmark}, two conclusions can be summarized.

\begin{table}[!htbp]
\centering
\caption{Results of merging Qwen1.5-Chat and Sailor on South-East Asian language benchmark.}
\label{tab:llms_merging_southeast_asian_benchmark}
\resizebox{1.0\textwidth}{!}
{
\setlength{\tabcolsep}{0.8mm}
{
\begin{tabular}{c|c|c|ccc|ccc|ccc|c|cc}
\hline
\multirow{2}{*}{Size} & \multirow{2}{*}{Models}                                                            & \multirow{2}{*}{\begin{tabular}[c]{@{}c@{}}Merging \\ Methods\end{tabular}} & XQuAD                & TydiQA                     & XQuAD                      & \multicolumn{3}{c|}{XCOPA}                       & \multicolumn{3}{c|}{Belebele}                    & M3Exam         & \multirow{2}{*}{Average} & \multirow{2}{*}{\begin{tabular}[c]{@{}c@{}}Average\\ Rank\end{tabular}} \\ \cline{4-13}
                      &                                                                                    &                                                                             & th                   & id                         & vi                         & th             & id             & vi             & th             & id             & vi             & jv             &                          &                                                                         \\ \hline
\multirow{9}{*}{7B}   & Qwen1.5                                                                            & /                                                                           & 53.79/69.30          & 57.17/77.28                & 56.63/76.99                & 54.20          & 62.20          & 66.20          & 38.33          & 42.00          & 42.89          & 26.15          & 55.63                    & /                                                                       \\
                      & Qwen1.5-Chat                                                                       & /                                                                           & 24.28/46.77          & 42.30/67.57                & 45.51/69.91                & 56.20          & 66.80          & 70.40          & 38.67          & 43.11          & 47.11          & 28.30          & 49.76                    & /                                                                       \\
                      & Sailor                                                                             & /                                                                           & 57.88/71.06          & 60.53/75.42                & 53.81/74.62                & 59.00          & 72.20          & 72.20          & 41.56          & 44.33          & 45.33          & 32.88          & 58.52                    & /                                                                       \\ \cline{2-15} 
                      & \multirow{6}{*}{\begin{tabular}[c]{@{}c@{}}Qwen1.5-Chat \\ \& Sailor\end{tabular}} & Task Arithmetic                                                             & {\underline{28.20}/49.62}    & {\textbf{45.84}/\underline{65.78}}          & 37.38/61.53                & \textbf{63.20} & \textbf{77.60} & {\underline{73.40}}    & {\underline{38.89}}    & 46.89          & 45.11          & {\underline{30.46}}    & {\underline{51.07}}              & {\underline{2.15}}                                                              \\
                      &                                                                                    & SLERP                                                                       & 16.62/43.62          & 20.53/54.02                & 33.70/61.49                & 55.80          & 73.40          & 73.00          & 38.44          & {\underline{47.89}}    & {\underline{47.56}}    & 28.30          & 45.72                    & 3.23                                                                    \\
                      &                                                                                    & Model Stock                                                                 & {26.72/\underline{52.69}}    & {\underline{24.78}/58.88}          & {\underline{43.80}/\underline{69.50}}          & 54.60          & 66.00          & 69.40          & 37.33          & 42.78          & 43.67          & 27.76          & 47.53                    & 3.31                                                                    \\
                      &                                                                                    & TIES-Merging                                                                & 0.61/8.84            & 5.66/17.23                 & 7.70/20.78                 & 50.20          & 62.20          & 59.80          & 30.22          & 35.33          & 35.11          & 25.07          & 27.60                    & 5.54                                                                    \\
                      &                                                                                    & Breadcrumbs                                                                 & 6.79/11.38           & 7.61/15.23                 & 12.32/27.90                & 51.40          & 66.40          & 57.20          & 31.33          & 34.00          & 32.56          & 24.53          & 29.13                    & 5.23                                                                    \\
                      &                                                                                    & WIDEN                                                                        & \textbf{42.65}/\textbf{64.21} & \textbf{45.84}/\textbf{73.37}       & \textbf{48.42}/\textbf{73.17}       & {\underline{60.20}}    & {\underline{77.40}}    & \textbf{73.60} & \textbf{40.11} & \textbf{51.11} & \textbf{48.56} & \textbf{32.88} & \textbf{56.27}           & \textbf{1.15}                                                           \\ \hline
\multirow{9}{*}{14B}  & Qwen1.5                                                                            & /                                                                           & 55.53/74.39          & 60.35/81.07                & 57.66/77.62                & 58.40          & 70.40          & 72.60          & 41.22          & 48.67          & 44.44          & 26.15          & 59.12                    & /                                                                       \\
                      & Qwen1.5-Chat                                                                       & /                                                                           & 33.59/59.98          & 37.17/65.46                & 44.14/71.91                & 61.80          & 75.20          & 71.80          & 44.00          & 51.00          & 52.67          & 29.92          & 53.74                    & /                                                                       \\
                      & Sailor                                                                             & /                                                                           & 49.43/70.01          & 58.94/77.85                & 57.74/77.34                & 62.60          & 77.60          & 78.60          & 40.89          & 47.67          & 47.11          & 32.88          & 59.90                    & /                                                                       \\ \cline{2-15} 
                      & \multirow{6}{*}{\begin{tabular}[c]{@{}c@{}}Qwen1.5-Chat \\ \& Sailor\end{tabular}} & Task Arithmetic                                                             & 8.53/24.39           & 13.45/33.54                & 13.52/25.75                & 59.80          & \textbf{82.40} & \textbf{78.20} & \textbf{46.00} & \textbf{56.33} & \textbf{53.78} & \textbf{33.69} & 40.72                    & 2.54                                                                    \\
                      &                                                                                    & SLERP                                                                       & 14.53/44.70          & {\underline{22.48}/\underline{61.67}}          & 42.69/69.48                & \textbf{61.80} & 75.60          & 74.60          & {\underline{43.22}}    & 52.56          & {\underline{50.56}}    & 29.92          & {\underline{49.52}}              & {\underline{2.46}}                                                              \\
                      &                                                                                    & Model Stock                                                                 & {\underline{25.59}/\underline{53.10}}    & 14.87/51.19                & {\underline{44.74}/\underline{70.20}}          & 58.60          & 70.40          & 71.80          & 42.67          & 49.89          & 45.11          & 27.22          & 48.11                    & 3.08                                                                    \\
                      &                                                                                    & TIES-Merging                                                                & 0.44/8.78            & 1.42/12.87                 & 0.00/6.95                  & 55.20          & 69.20          & 67.20          & 32.78          & 39.00          & 37.11          & 27.22          & 27.55                    & 5.46                                                                    \\
                      &                                                                                    & Breadcrumbs                                                                 & 1.22/6.48            & 2.30/20.88                 & 3.17/14.46                 & 52.20          & 64.60          & 63.40          & 34.78          & 42.11          & 40.67          & 26.68          & 28.69                    & 5.23                                                                    \\
                      &                                                                                    & WIDEN                                                                        & \textbf{49.61}/\textbf{73.16} & \textbf{50.62}/\textbf{75.09}       & \textbf{54.75}/\textbf{78.23}       & {\underline{60.80}}    & {\underline{77.40}}    & {\underline{74.60}}    & 42.22          & {\underline{56.22}}    & 50.44          & {\underline{32.61}}    & \textbf{59.67}           & \textbf{1.77}                                                           \\ \hline
\end{tabular}
}
}
\end{table}

Firstly, \textit{existing model merging approaches encounter significant challenges when incorporating the multilingual abilities of Sailor, leading to a marked decline in performance}. The downturn is probably attributed to the difficulty in determining the optimal combination due to diverse parameter changed ranges between Qwen1.5-Chat and Sailor. We also notice that the reduction is particularly pronounced in pruning-based methods, prompting us to conduct additional verifications. As demonstrated in \tabref{tab:sailor_7b_pruning_results}, we find that the feasibility of pruning strategies such as DARE and Magnitude-based Pruning (MP) in TIES-Merging and Breadcrumbs is severely compromised with minor parameter drop rates on Sailor-7B, far below the levels reported results in the original studies (i.e., 0.9 in DARE, 0.8 in TIES-Merging, and 0.85 in Breadcrumbs), diminishing the effectiveness of pruning in alleviating parameter interference. As a result, DARE fails to serve as a plug-in for existing merging techniques when considering PT LLMs, and its inferior results are excluded.

\begin{wraptable}{r}{0.55\textwidth}
\centering
\vspace{-18pt}
\caption{Performance of pruning strategies on Sailor-7B for Vietnamese-related tasks.}
\label{tab:sailor_7b_pruning_results}
\setlength{\tabcolsep}{0.8mm}
{
\begin{tabular}{c|cccc}
\hline
                      & Drop Rate & XQuAD  & XCOPA & Belebele \\ \hline
Sailor-7B             & /         & 53.81/74.62 & 72.20      & 45.33         \\ \hline
\multirow{2}{*}{DARE} & 0.1       & 47.56/66.95 & 64.20      & 41.00         \\
                      & 0.3       & 5.90/16.05  & 55.60      & 30.56         \\ \hline
\multirow{4}{*}{MP}   & 0.1       & 54.23/75.16 & 72.80      & 45.44         \\
                      & 0.3       & 52.44/73.53 & 72.20      & 44.78         \\
                      & 0.5       & 49.19/70.11 & 70.00      & 43.67         \\
                      & 0.8       & 13.77/30.13 & 59.00      & 34.56         \\ \hline
\end{tabular}
}
\vspace{-20pt}
\end{wraptable}

Secondly, \textit{WIDEN effectively assimilates the multilingual capabilities of Sailor, emerging as the top performer among all the merging techniques}. The key advantage of WIDEN lies in the adaptive computation of weight importance by considering both magnitudes and directions during the merging process, mitigating the influence of diverse parameter changed ranges between FT and PT LLMs.

\begin{table}[!htbp]
\centering
\caption{Performance of merging Qwen1.5-Chat and Sailor on Open LLM Leaderboard.}
\label{tab:llms_merging_open_llm_leaderboard}
\resizebox{1.0\textwidth}{!}
{
\setlength{\tabcolsep}{1.0mm}
{
\begin{tabular}{c|c|c|cccccc|cc}
\hline
Size                  & Models                                                                             & \begin{tabular}[c]{@{}c@{}}Merging \\ Methods\end{tabular} & ARC            & \begin{tabular}[c]{@{}c@{}}Hella-\\ Swag\end{tabular} & MMLU           & \begin{tabular}[c]{@{}c@{}}Truthful-\\ QA\end{tabular} & \begin{tabular}[c]{@{}c@{}}Wino-\\ grande\end{tabular} & GSM8K          & Average        & \begin{tabular}[c]{@{}c@{}}Average \\ Rank\end{tabular} \\ \hline
\multirow{9}{*}{7B}   & Qwen1.5                                                                            & /                                                          & 54.86          & 78.45                                                 & 60.60          & 51.09                                                  & 71.03                                                  & 56.79          & 62.14          & /                                                       \\
                      & Qwen1.5-Chat                                                                       & /                                                          & 56.14          & 78.71                                                 & 60.18          & 53.61                                                  & 67.48                                                  & 54.21          & 61.72          & /                                                       \\
                      & Sailor                                                                             & /                                                          & 49.57          & 76.13                                                 & 52.91          & 40.07                                                  & 71.35                                                  & 34.65          & 54.11          & /                                                       \\ \cline{2-11} 
                      & \multirow{6}{*}{\begin{tabular}[c]{@{}c@{}}Qwen1.5-Chat \\ \& Sailor\end{tabular}} & Task Arithmetic                                            & 52.05          & 75.15                                                 & 59.38          & {\underline{50.84}}                                            & 69.77                                                  & 25.55          & 55.46          & 3.50                                                    \\
                      &                                                                                    & SLERP                                                      & {\underline{54.78}}    & 76.20                                                 & {\underline{60.76}}    & 50.78                                                  & {\underline{71.51}}                                            & {\underline{55.50}}    & \textbf{61.59} & {\underline{2.33}}                                              \\
                      &                                                                                    & Model Stock                                                & \textbf{55.12} & \textbf{76.29}                                        & \textbf{61.18} & 49.33                                                  & 71.43                                                  & \textbf{55.80} & {\underline{61.53}}    & \textbf{2.00}                                           \\
                      &                                                                                    & TIES-Merging                                               & 43.86          & 56.88                                                 & 52.39          & 46.59                                                  & 67.56                                                  & 0.00           & 44.55          & 5.67                                                    \\
                      &                                                                                    & Breadcrumbs                                                & 47.18          & 49.99                                                 & 52.66          & \textbf{52.05}                                         & 64.88                                                  & 0.45           & 44.53          & 4.67                                                    \\
                      &                                                                                    & WIDEN                                                       & 53.84          & {\underline{76.25}}                                           & 57.65          & 49.34                                                  & \textbf{71.90}                                         & 44.81          & 58.97          & 2.83                                                    \\ \hline
\multirow{9}{*}{14B}  & Qwen1.5                                                                            & /                                                          & 56.40          & 81.22                                                 & 67.79          & 52.04                                                  & 74.43                                                  & 68.01          & 66.65          & /                                                       \\
                      & Qwen1.5-Chat                                                                       & /                                                          & 57.25          & 82.56                                                 & 67.48          & 60.42                                                  & 72.69                                                  & 68.08          & 68.08          & /                                                       \\
                      & Sailor                                                                             & /                                                          & 55.46          & 80.31                                                 & 62.95          & 46.64                                                  & 76.80                                                  & 61.94          & 64.02          & /                                                       \\ \cline{2-11} 
                      & \multirow{6}{*}{\begin{tabular}[c]{@{}c@{}}Qwen1.5-Chat \\ \& Sailor\end{tabular}} & Task Arithmetic                                            & 56.57          & \textbf{81.59}                                        & 67.52          & \textbf{62.93}                                         & 75.22                                                  & 53.98          & 66.30          & {\underline{2.50}}                                              \\
                      &                                                                                    & SLERP                                                      & 55.72          & 79.94                                                 & {\underline{67.94}}    & 57.51                                                  & 75.14                                                  & \textbf{69.29} & \textbf{67.59} & 3.00                                                    \\
                      &                                                                                    & Model Stock                                                & {\underline{57.00}}    & {\underline{80.50}}                                           & \textbf{68.44} & 51.98                                                  & {\underline{76.01}}                                            & {\underline{66.72}}    & {\underline{66.77}}    & \textbf{2.33}                                           \\
                      &                                                                                    & TIES-Merging                                               & 49.74          & 67.23                                                 & 60.54          & 47.43                                                  & 72.14                                                  & 0.30           & 49.56          & 5.67                                                    \\
                      &                                                                                    & Breadcrumbs                                                & 51.88          & 62.22                                                 & 63.47          & {\underline{57.90}}                                            & 70.32                                                  & 4.55           & 51.72          & 4.83                                                    \\
                      &                                                                                    & WIDEN                                                       & \textbf{57.17} & 80.05                                                 & 66.00          & 54.85                                                  & \textbf{76.09}                                         & 66.34          & 66.75          & 2.67                                                    \\ \hline
\end{tabular}
}
}
\end{table}

\tabref{tab:llms_merging_open_llm_leaderboard} depicts the merging performance on Open LLM Leaderboard. We find that geometric-based approaches (SLERP and Model Stock) excel in retraining the instruction-following skills of Qwen1.5-Chat, indicating that parameters of FT LLMs may potentially exhibit more evident properties in the geometric space. WIDEN shows competitive results alongside SLERP and Model Stock, underscoring its applicability in merging FT LLMs. Moreover, WIDEN outperforms arithmetic-based methods since it is a generalized format of these methods and offers greater flexibility through the adaptive computation of weight importance.

\subsection{Performance of Merging FT LLMs}
Under the setting of merging multiple FT LLMs, we strictly follow the identical protocol in \citet{yu2023language} and report the official results in \tabref{tab:llms_merging_instruct_math_code} for fair comparisons. One exception is that we use AlpacaEval 2.0 instead of AlpacaEval in \citet{yu2023language} for evaluation, aiming to provide more convincing and reliable verifications. We evaluate the performance of merging pairwise LLMs since some baselines (e.g., SLERP) are only applicable for dealing with two models. 

\begin{table}[!htbp]
\centering
\caption{Performance of merging WizardLM-13B, WizardMath-13B, and llama-2-13b-code-alpaca.}
\label{tab:llms_merging_instruct_math_code}
\resizebox{0.95\textwidth}{!}
{
\setlength{\tabcolsep}{1.0mm}
{
\begin{tabular}{c|c|c|cc|cc}
\hline
\multirow{2}{*}{Models}                                                           & \multirow{2}{*}{Merging Methods} & \begin{tabular}[c]{@{}c@{}}Instruction-\\ following\end{tabular} & \multicolumn{2}{c|}{\begin{tabular}[c]{@{}c@{}}Mathematical \\ Reasoning\end{tabular}} & \multicolumn{2}{c}{Code Generation} \\ \cline{3-7} 
                                                                                  &                                  & AlpacaEval 2.0                                                       & GSM8K                                      & MATH                                      & HumanEval          & MBPP           \\ \hline
WizardLM-13B                                                                                & /                                & 12.73                                                           & 2.20                                       & 0.04                                      & 36.59              & 34.00          \\ \hline
WizardMath-13B                                                                             & /                                & /                                                                & 64.22                                      & 14.02                                     & /                  & /              \\ \hline
llama-2-13b-code-alpaca                                                                              & /                                & /                                                                & /                                          & /                                         & 23.78              & 27.60          \\ \hline
\multirow{6}{*}{\begin{tabular}[c]{@{}c@{}}WizardLM-13B \\ \& WizardMath-13B\end{tabular}}            & Task Arithmetic                  &       \textbf{11.85}                                                     & \textbf{66.34}                                      & 13.40                                     & \underline{28.66}              & 30.60          \\
                                                                                  & SLERP                            &      7.90                                                            & \underline{66.19}                                      & \underline{13.44}                                     & 28.05              & 30.80          \\
                                                                                  & Model Stock                      &              0.25                                                    & 0.00                                       & 0.00                                      & 3.05               & 25.80          \\
                                                                                  & TIES-Merging                     & \underline{10.07}                                                            & 15.77                                      & 2.04                                      & \textbf{37.80}              & \textbf{35.60}          \\                                                                                  & Breadcrumbs                      &      9.85                                                            & 64.75                                      & 11.80                                     & 26.22              & \underline{33.20}          \\
                                                                                  & WIDEN                             &              9.45                                                    & \textbf{66.34}                                      & \textbf{13.58}                                     & \underline{28.66}              & 30.40          \\ \hline
\multirow{6}{*}{\begin{tabular}[c]{@{}c@{}}WizardLM-13B \\ \& llama-2-13b-code-alpaca \end{tabular}}            & Task Arithmetic                  & \textbf{10.09}                                                            & /                                          & /                                         & 31.70              & 32.40          \\
                                                                                  & SLERP                            &                    6.04                                              & /                                          & /                                         & \underline{32.32}              & \textbf{35.80}          \\
                                                                                  & Model Stock                      &                0.25                                                  & /                                          & /                                         & 3.66               & 24.80          \\
                                                                                  & TIES-Merging                     & \underline{7.27}                                                            & /                                          & /                                         & 0.00               & 0.00           \\
                                                                                  & Breadcrumbs                      &             7.23                                                     & /                                          & /                                         & \textbf{33.54}              & 32.00          \\
                                                                                  & WIDEN                             &               6.53                                                   & /                                          & /                                         & 31.70              & \underline{35.60}          \\ \hline
\multirow{6}{*}{\begin{tabular}[c]{@{}c@{}}WizardMath-13B\\ \& llama-2-13b-code-alpaca \end{tabular}}          & Task Arithmetic                  & /                                                                & \textbf{64.67}                                      & \textbf{13.98}                                     & 8.54               & 8.60           \\
                                                                                  & SLERP                            & /                                                                & 61.41                                      & 12.50                                     & \underline{9.15}               & \underline{22.40}          \\
                                                                                  & Model Stock                      & /                                                                & 0.00                                       & 0.00                                      & 4.27               & \textbf{25.60}          \\
                                                                                  & TIES-Merging                     & /                                                                & 63.23                                      & 13.56                                     & \textbf{9.76}               & \underline{22.40}          \\
                                                                                  & Breadcrumbs                      & /                                                                & 62.55                                      & 12.48                                     & \underline{9.15}               & 16.20          \\
                                                                                  & WIDEN                             & /                                                                & \underline{64.22}                                      & \underline{13.58}                                     & \textbf{9.76}               & 9.80           \\ \hline
\end{tabular}
}
}
\end{table}

From \tabref{tab:llms_merging_instruct_math_code}, we observe that the efficacy of certain baselines drastically fluctuates when integrating FT LLMs. For example, Model Stock appears to lose potency, whereas pruning-based methods including TIES-Merging and Breadcrumbs show competitive performance. WIDEN consistently depicts results on par with established merging techniques in most situations, affirming its suitability in the standard setting of merging multiple FT LLMs.

\subsection{Investigations of Designs in WIDEN}
\begin{wrapfigure}{r}{0.5\textwidth}
    \centering
    \vspace{-12pt}
    \includegraphics[width=0.5\columnwidth]{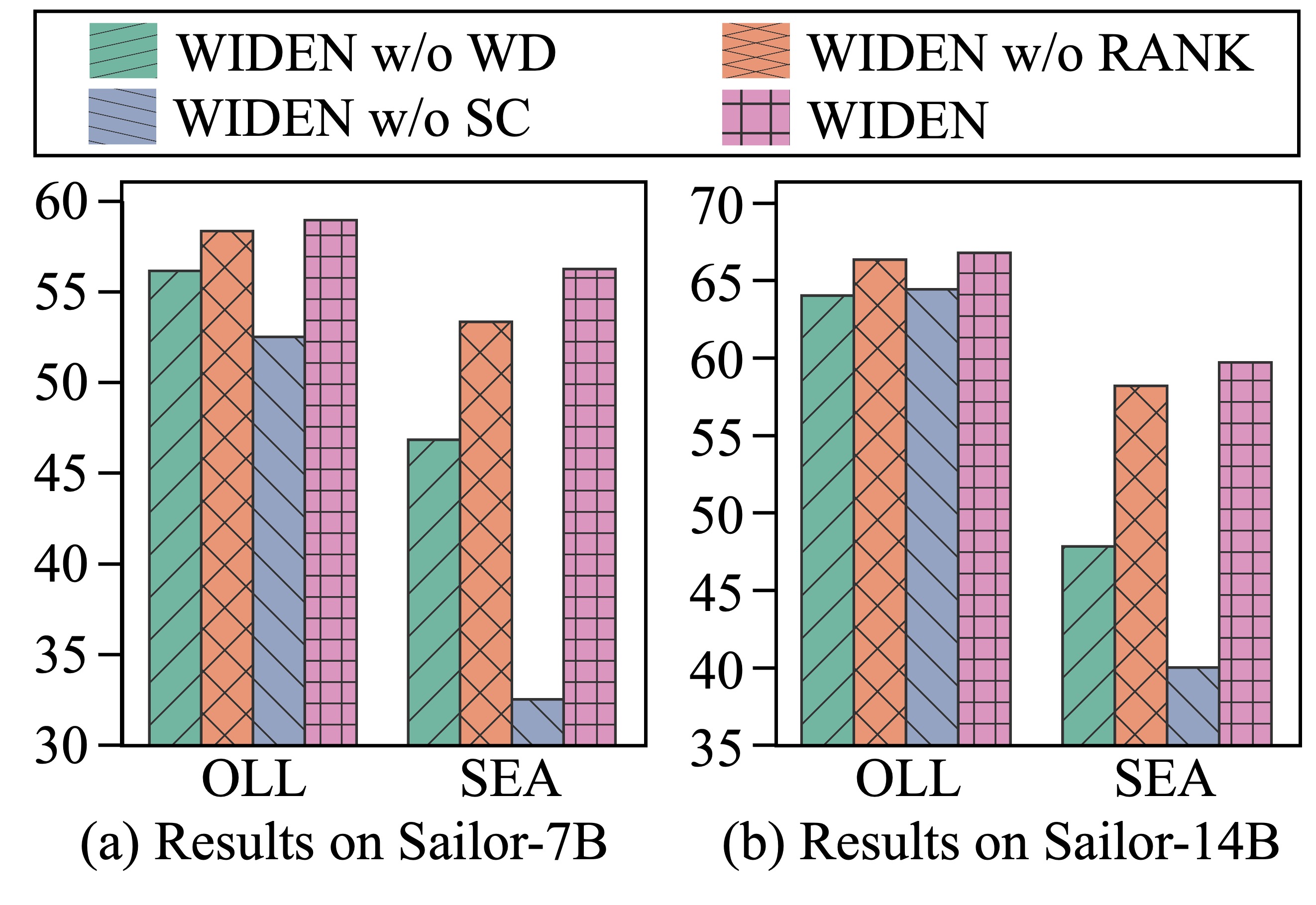}
    \caption{Effects of various designs in WIDEN.}
    \label{fig:ablation_study}
    \vspace{-12pt}
\end{wrapfigure}

The foundational designs in WIDEN consist of three components: weight disentanglement, ranking weights inside each model, and score calibration for Softmax. To assess the contribution of each module, we respectively remove the above components and measure the performance of the remaining parts. Specifically, we eliminate the disentanglement of weights by calculating the discrepancy between the weights of LLM and the corresponding backbone using cosine similarities, denoted as WIDEN w/o WD. We substitute the ranking mechanism with min-max normalization within each model, represented by WIDEN w/o RANK. We discard the score calibration and directly employ Softmax to compute importance scores, identified as WIDEN w/o SC. \figref{fig:ablation_study} shows the impact of these three modifications, where OLL and SEA are the abbreviations for Open LLM Leaderboard and South-East Asian language benchmark, respectively. Note that the reported results are the average of metrics across all the datasets within each benchmark.

From \figref{fig:ablation_study}, we find that each design in WIDEN contributes to enhancing the merging performance, particularly in absorbing the multilingual abilities on the South-East Asian language benchmark. Precisely, the weight disentanglement refines the estimation of weight importance at a granular level, considering both magnitude and direction. The ranking mechanism offers a smoother distribution of weight importance based on continuous indices, effectively mitigating the influence of diverse parameter changed ranges. The calibration of scores computed by Softmax reallocates importance to critical parameters, which maintains the characteristics of essential parameters across multiple models. In summary, the components of WIDEN are indispensable and improve performance with varied benefits; the removal of any module leads to diminished outcomes.

\subsection{Analysis of Computed Weight Importance}
\begin{figure}[!htbp]
    \centering
    \includegraphics[width=1.0\columnwidth]{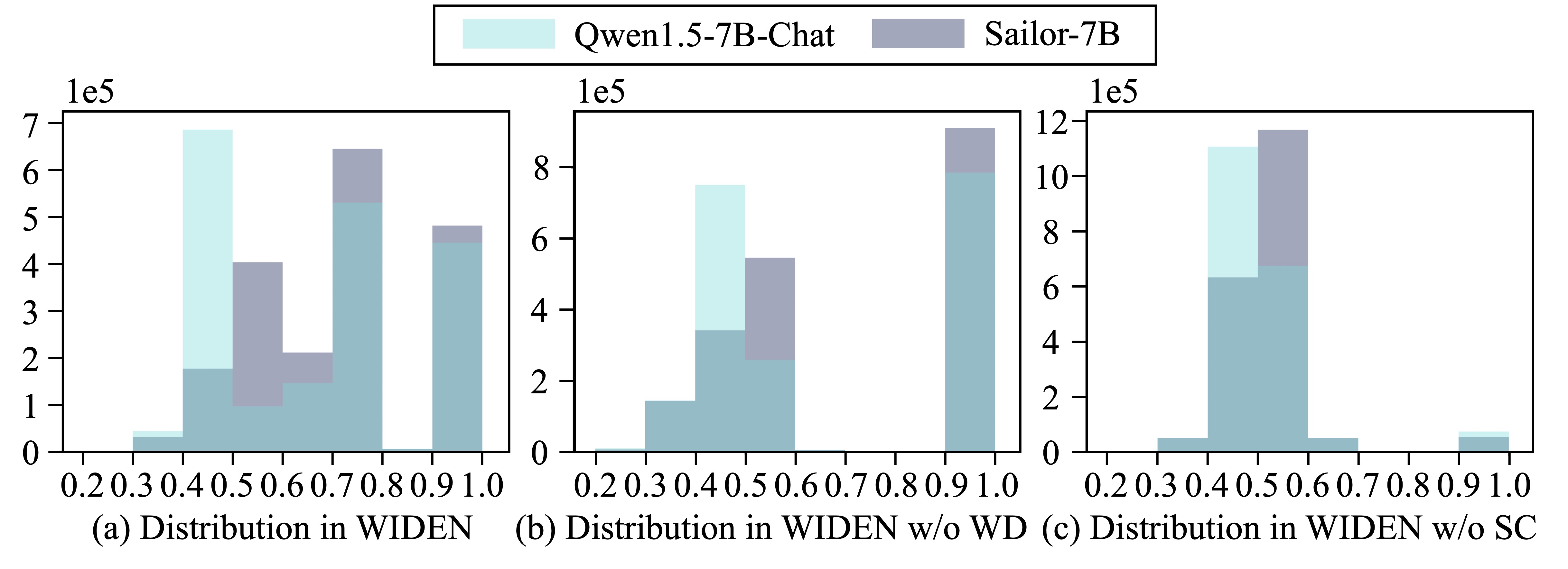}
    \caption{Distribution of weight importance computed by WIDEN and its variations.}
    \label{fig:computed_weight_importance}
\end{figure}

We further delve into the properties of weight importance calculated by WIDEN from both qualitative and quantitative perspectives. 
Since \figref{fig:ablation_study} demonstrates that the improvements in weight disentanglement and score calibration are notably more pronounced, we qualitatively depict the distribution of weight importance computed by WIDEN, WIDEN w/o WD, and WIDEN w/o SC on 7B model size in \figref{fig:computed_weight_importance}. Our observations reveal that: 1) WIDEN exhibits a more balanced and reasonable weight importance distribution than WIDEN w/o WD, attributed to the disentanglement of weights. The distribution of WIDEN ranges approximately from 0.3 to 0.8 and 0.9 to 1.0, versus 0.3 to 0.6 and 0.9 to 1.0 for WIDEN w/o WD. WIDEN considers the collective contributions of magnitude and direction, rather than the individual impacts of weights, leading to a more holistic assessment of weight importance with increased numbers of weights falling within the importance range from 0.6 to 0.8. As a result, compared with WIDEN w/o WD, WIDEN achieves 4.98\% and 20.08\% improvements on average on the Open LLM Leaderboard and the South-East Asian language benchmark, respectively; 2) In contrast to WIDEN w/o SC, WIDEN distinguishes essential weights and assigns high importance within the range of 0.6 to 0.8 as well as 0.9 to 1.0 for certain weights, thanks to the design of score calibration. Therefore, WIDEN ensures the retention of essential weights in both Qwen1.5-7B-Chat and Sailor-7B, resulting in 12.25\% and 72.87\% average enhancements on the two benchmarks.

Furthermore, we categorize weight importance into three levels: Low (L), Medium (M), and High (H). The Low tier comprises the first third of weights when sorted by ascending importance, indicating those with the least significance. The Medium tier includes weights from the 1/3 mark to the 2/3 mark, and the High tier contains weights from the 2/3 mark to the end. \tabref{tab:weight_importance_transfer} quantitatively illustrates the adjustments of weight importance made by WIDEN when compared to WIDEN w/o WD and WIDEN w/o SC across three levels. We find that WIDEN effectively reallocates the weight importance via three aspects: 1) elevating weights of lower importance from Low to Medium; 2) either demoting or promoting weights of medium importance from Medium to Low or from Medium to High, respectively; 3) decreasing weights of high importance from High to Medium. These adjustments in weight importance explain how WIDEN brings improvements through the designs of weight disentanglement and score calibration.
\begin{table}[!htbp]
\centering
\caption{Adjustments of weight importance made by WIDEN.}
\label{tab:weight_importance_transfer}
\resizebox{1.0\textwidth}{!}
{
\setlength{\tabcolsep}{0.7mm}
{
\begin{tabular}{c|c|ccccccccc}
\hline
Adjustments                                                                       & Models          & L$\rightarrow$L    & L$\rightarrow$M    & L$\rightarrow$H    & M$\rightarrow$L    & M$\rightarrow$M    & M$\rightarrow$H    & H$\rightarrow$L    & H$\rightarrow$M    & H$\rightarrow$H    \\ \hline
\multirow{2}{*}{\begin{tabular}[c]{@{}c@{}}WIDEN w/o WD\\ to WIDEN\end{tabular}} & Qwen1.5-7B-Chat & 18.82\% & \textcolor{red!60!white}{11.09\%} & 3.42\% & \textcolor{red!60!white}{13.97\%} & 10.18\% & \textcolor{red!60!white}{9.18\%} & 0.54\% & \textcolor{red!60!white}{12.06\%} & 20.75\% \\
                                                                                 & Sailor-7B       & 15.34\% & \textcolor{red!60!white}{10.50\%} & \textcolor{red!60!white}{7.48\%} & \textcolor{red!60!white}{17.80\%} & 7.72\% & \textcolor{red!60!white}{7.80\%} & 0.18\% & \textcolor{red!60!white}{15.10\%} & 18.07\% \\ \hline
\multirow{2}{*}{\begin{tabular}[c]{@{}c@{}}WIDEN w/o SC\\ to WIDEN\end{tabular}} & Qwen1.5-7B-Chat & 24.78\% & \textcolor{red!60!white}{7.69\%} & 0.85\% & \textcolor{red!60!white}{7.93\%} & 17.51\% & \textcolor{red!60!white}{7.88\%} & 0.62\% & \textcolor{red!60!white}{8.12\%} & 24.61\% \\
                                                                                 & Sailor-7B       & 22.01\% & \textcolor{red!60!white}{9.52\%} & 1.80\% & \textcolor{red!60!white}{9.63\%} & 15.14\% & \textcolor{red!60!white}{8.56\%} & 1.69\% & \textcolor{red!60!white}{8.67\%} & 22.99\% \\ \hline
\end{tabular}
}
}
\end{table}

\section{Conclusion}
\label{section-5}
In this study, we paved the way for extending the merging scope from FT to PT LLMs. Specifically, we first observed that existing methods struggled to integrate the abilities of PT LLMs and then introduced WIDEN, an innovative approach based on weight disentanglement, to effectively deploy merging strategies to PT LLMs. Experimental findings demonstrated that WIDEN not only exhibited an advantage in absorbing the abilities of PT LLMs but also preserved the skills of FT LLMs. Additionally, WIDEN achieved competitive performance with established merging methods in the conventional setting of merging FT LLMs. We further offered a detailed analysis of the designs underlying WIDEN. This work made the first attempt to broaden the sources of combinable abilities, fostering the broader application of model merging techniques.


\bibliography{reference.bib}

\begin{thebibliography}{56}
\providecommand{\natexlab}[1]{#1}
\providecommand{\url}[1]{\texttt{#1}}
\expandafter\ifx\csname urlstyle\endcsname\relax
  \providecommand{\doi}[1]{doi: #1}\else
  \providecommand{\doi}{doi: \begingroup \urlstyle{rm}\Url}\fi

\bibitem[Artetxe et~al.(2020)Artetxe, Ruder, and Yogatama]{DBLP:conf/acl/ArtetxeRY20}
Mikel Artetxe, Sebastian Ruder, and Dani Yogatama.
\newblock On the cross-lingual transferability of monolingual representations.
\newblock In \emph{Proceedings of the 58th Annual Meeting of the Association for Computational Linguistics}, pp.\  4623--4637. Association for Computational Linguistics, 2020.

\bibitem[Austin et~al.(2021)Austin, Odena, Nye, Bosma, Michalewski, Dohan, Jiang, Cai, Terry, Le, and Sutton]{DBLP:journals/corr/abs-2108-07732}
Jacob Austin, Augustus Odena, Maxwell~I. Nye, Maarten Bosma, Henryk Michalewski, David Dohan, Ellen Jiang, Carrie~J. Cai, Michael Terry, Quoc~V. Le, and Charles Sutton.
\newblock Program synthesis with large language models.
\newblock \emph{CoRR}, abs/2108.07732, 2021.

\bibitem[Bai et~al.(2023)Bai, Bai, Chu, Cui, Dang, Deng, Fan, Ge, Han, Huang, Hui, Ji, Li, Lin, Lin, Liu, Liu, Lu, Lu, Ma, Men, Ren, Ren, Tan, Tan, Tu, Wang, Wang, Wang, Wu, Xu, Xu, Yang, Yang, Yang, Yang, Yao, Yu, Yuan, Yuan, Zhang, Zhang, Zhang, Zhang, Zhou, Zhou, Zhou, and Zhu]{DBLP:journals/corr/abs-2309-16609}
Jinze Bai, Shuai Bai, Yunfei Chu, Zeyu Cui, Kai Dang, Xiaodong Deng, Yang Fan, Wenbin Ge, Yu~Han, Fei Huang, Binyuan Hui, Luo Ji, Mei Li, Junyang Lin, Runji Lin, Dayiheng Liu, Gao Liu, Chengqiang Lu, Keming Lu, Jianxin Ma, Rui Men, Xingzhang Ren, Xuancheng Ren, Chuanqi Tan, Sinan Tan, Jianhong Tu, Peng Wang, Shijie Wang, Wei Wang, Shengguang Wu, Benfeng Xu, Jin Xu, An~Yang, Hao Yang, Jian Yang, Shusheng Yang, Yang Yao, Bowen Yu, Hongyi Yuan, Zheng Yuan, Jianwei Zhang, Xingxuan Zhang, Yichang Zhang, Zhenru Zhang, Chang Zhou, Jingren Zhou, Xiaohuan Zhou, and Tianhang Zhu.
\newblock Qwen technical report.
\newblock \emph{CoRR}, abs/2309.16609, 2023.

\bibitem[Bandarkar et~al.(2023)Bandarkar, Liang, Muller, Artetxe, Shukla, Husa, Goyal, Krishnan, Zettlemoyer, and Khabsa]{DBLP:journals/corr/abs-2308-16884}
Lucas Bandarkar, Davis Liang, Benjamin Muller, Mikel Artetxe, Satya~Narayan Shukla, Donald Husa, Naman Goyal, Abhinandan Krishnan, Luke Zettlemoyer, and Madian Khabsa.
\newblock The belebele benchmark: a parallel reading comprehension dataset in 122 language variants.
\newblock \emph{CoRR}, abs/2308.16884, 2023.

\bibitem[Beeching et~al.(2023)Beeching, Fourrier, Habib, Han, Lambert, Rajani, Sanseviero, Tunstall, and Wolf]{open-llm-leaderboard}
Edward Beeching, Clémentine Fourrier, Nathan Habib, Sheon Han, Nathan Lambert, Nazneen Rajani, Omar Sanseviero, Lewis Tunstall, and Thomas Wolf.
\newblock Open llm leaderboard, 2023.

\bibitem[Chaudhary(2023)]{codealpaca}
Sahil Chaudhary.
\newblock Code alpaca: An instruction-following llama model for code generation.
\newblock \url{https://github.com/sahil280114/codealpaca}, 2023.

\bibitem[Chen et~al.(2021)Chen, Tworek, Jun, Yuan, de~Oliveira~Pinto, Kaplan, Edwards, Burda, Joseph, Brockman, Ray, Puri, Krueger, Petrov, Khlaaf, Sastry, Mishkin, Chan, Gray, Ryder, Pavlov, Power, Kaiser, Bavarian, Winter, Tillet, Such, Cummings, Plappert, Chantzis, Barnes, Herbert{-}Voss, Guss, Nichol, Paino, Tezak, Tang, Babuschkin, Balaji, Jain, Saunders, Hesse, Carr, Leike, Achiam, Misra, Morikawa, Radford, Knight, Brundage, Murati, Mayer, Welinder, McGrew, Amodei, McCandlish, Sutskever, and Zaremba]{DBLP:journals/corr/abs-2107-03374}
Mark Chen, Jerry Tworek, Heewoo Jun, Qiming Yuan, Henrique~Pond{\'{e}} de~Oliveira~Pinto, Jared Kaplan, Harrison Edwards, Yuri Burda, Nicholas Joseph, Greg Brockman, Alex Ray, Raul Puri, Gretchen Krueger, Michael Petrov, Heidy Khlaaf, Girish Sastry, Pamela Mishkin, Brooke Chan, Scott Gray, Nick Ryder, Mikhail Pavlov, Alethea Power, Lukasz Kaiser, Mohammad Bavarian, Clemens Winter, Philippe Tillet, Felipe~Petroski Such, Dave Cummings, Matthias Plappert, Fotios Chantzis, Elizabeth Barnes, Ariel Herbert{-}Voss, William~Hebgen Guss, Alex Nichol, Alex Paino, Nikolas Tezak, Jie Tang, Igor Babuschkin, Suchir Balaji, Shantanu Jain, William Saunders, Christopher Hesse, Andrew~N. Carr, Jan Leike, Joshua Achiam, Vedant Misra, Evan Morikawa, Alec Radford, Matthew Knight, Miles Brundage, Mira Murati, Katie Mayer, Peter Welinder, Bob McGrew, Dario Amodei, Sam McCandlish, Ilya Sutskever, and Wojciech Zaremba.
\newblock Evaluating large language models trained on code.
\newblock \emph{CoRR}, abs/2107.03374, 2021.

\bibitem[Cheng et~al.(2024)Cheng, Huang, and Wei]{DBLP:journals/corr/abs-2309-09530}
Daixuan Cheng, Shaohan Huang, and Furu Wei.
\newblock Adapting large language models via reading comprehension.
\newblock In \emph{The Twelfth International Conference on Learning Representations}. OpenReview.net, 2024.

\bibitem[Clark et~al.(2020)Clark, Palomaki, Nikolaev, Choi, Garrette, Collins, and Kwiatkowski]{DBLP:journals/tacl/ClarkPNCGCK20}
Jonathan~H. Clark, Jennimaria Palomaki, Vitaly Nikolaev, Eunsol Choi, Dan Garrette, Michael Collins, and Tom Kwiatkowski.
\newblock Tydi {QA:} {A} benchmark for information-seeking question answering in typologically diverse languages.
\newblock \emph{Trans. Assoc. Comput. Linguistics}, 8:\penalty0 454--470, 2020.

\bibitem[Clark et~al.(2018)Clark, Cowhey, Etzioni, Khot, Sabharwal, Schoenick, and Tafjord]{DBLP:journals/corr/abs-1803-05457}
Peter Clark, Isaac Cowhey, Oren Etzioni, Tushar Khot, Ashish Sabharwal, Carissa Schoenick, and Oyvind Tafjord.
\newblock Think you have solved question answering? try arc, the {AI2} reasoning challenge.
\newblock \emph{CoRR}, abs/1803.05457, 2018.

\bibitem[Cobbe et~al.(2021)Cobbe, Kosaraju, Bavarian, Chen, Jun, Kaiser, Plappert, Tworek, Hilton, Nakano, Hesse, and Schulman]{DBLP:journals/corr/abs-2110-14168}
Karl Cobbe, Vineet Kosaraju, Mohammad Bavarian, Mark Chen, Heewoo Jun, Lukasz Kaiser, Matthias Plappert, Jerry Tworek, Jacob Hilton, Reiichiro Nakano, Christopher Hesse, and John Schulman.
\newblock Training verifiers to solve math word problems.
\newblock \emph{CoRR}, abs/2110.14168, 2021.

\bibitem[Colombo et~al.(2024)Colombo, Pires, Boudiaf, Culver, Melo, Corro, Martins, Esposito, Raposo, Morgado, and Desa]{colombo2024saullm}
Pierre Colombo, Telmo~Pessoa Pires, Malik Boudiaf, Dominic Culver, Rui Melo, Caio Corro, Andre~FT Martins, Fabrizio Esposito, Vera~L{\'u}cia Raposo, Sofia Morgado, and Michael Desa.
\newblock Saullm-7b: A pioneering large language model for law.
\newblock \emph{arXiv preprint arXiv:2403.03883}, 2024.

\bibitem[Crawshaw(2020)]{DBLP:journals/corr/abs-2009-09796}
Michael Crawshaw.
\newblock Multi-task learning with deep neural networks: {A} survey.
\newblock \emph{CoRR}, abs/2009.09796, 2020.

\bibitem[Davari \& Belilovsky(2023)Davari and Belilovsky]{DBLP:journals/corr/abs-2312-06795}
MohammadReza Davari and Eugene Belilovsky.
\newblock Model breadcrumbs: Scaling multi-task model merging with sparse masks.
\newblock \emph{CoRR}, abs/2312.06795, 2023.

\bibitem[Devlin et~al.(2019)Devlin, Chang, Lee, and Toutanova]{DBLP:conf/naacl/DevlinCLT19}
Jacob Devlin, Ming{-}Wei Chang, Kenton Lee, and Kristina Toutanova.
\newblock {BERT:} pre-training of deep bidirectional transformers for language understanding.
\newblock In \emph{Proceedings of the 2019 Conference of the North American Chapter of the Association for Computational Linguistics: Human Language Technologies}, pp.\  4171--4186. Association for Computational Linguistics, 2019.

\bibitem[Dou et~al.(2024)Dou, Liu, Zeng, Guo, Zhou, Lu, and Lin]{DBLP:journals/corr/abs-2404-03608}
Longxu Dou, Qian Liu, Guangtao Zeng, Jia Guo, Jiahui Zhou, Wei Lu, and Min Lin.
\newblock Sailor: Open language models for south-east asia.
\newblock \emph{CoRR}, abs/2404.03608, 2024.

\bibitem[Dubois et~al.(2024)Dubois, Galambosi, Liang, and Hashimoto]{DBLP:journals/corr/abs-2404-04475}
Yann Dubois, Bal{\'{a}}zs Galambosi, Percy Liang, and Tatsunori~B. Hashimoto.
\newblock Length-controlled alpacaeval: {A} simple way to debias automatic evaluators.
\newblock \emph{CoRR}, abs/2404.04475, 2024.

\bibitem[Gao et~al.(2023)Gao, Tow, Abbasi, Biderman, Black, DiPofi, Foster, Golding, Hsu, Le~Noac'h, Li, McDonell, Muennighoff, Ociepa, Phang, Reynolds, Schoelkopf, Skowron, Sutawika, Tang, Thite, Wang, Wang, and Zou]{eval-harness}
Leo Gao, Jonathan Tow, Baber Abbasi, Stella Biderman, Sid Black, Anthony DiPofi, Charles Foster, Laurence Golding, Jeffrey Hsu, Alain Le~Noac'h, Haonan Li, Kyle McDonell, Niklas Muennighoff, Chris Ociepa, Jason Phang, Laria Reynolds, Hailey Schoelkopf, Aviya Skowron, Lintang Sutawika, Eric Tang, Anish Thite, Ben Wang, Kevin Wang, and Andy Zou.
\newblock A framework for few-shot language model evaluation, 12 2023.

\bibitem[Goddard et~al.(2024)Goddard, Siriwardhana, Ehghaghi, Meyers, Karpukhin, Benedict, McQuade, and Solawetz]{DBLP:journals/corr/abs-2403-13257}
Charles Goddard, Shamane Siriwardhana, Malikeh Ehghaghi, Luke Meyers, Vlad Karpukhin, Brian Benedict, Mark McQuade, and Jacob Solawetz.
\newblock Arcee's mergekit: {A} toolkit for merging large language models.
\newblock \emph{CoRR}, abs/2403.13257, 2024.

\bibitem[Hendrycks et~al.(2021{\natexlab{a}})Hendrycks, Burns, Basart, Zou, Mazeika, Song, and Steinhardt]{DBLP:conf/iclr/HendrycksBBZMSS21}
Dan Hendrycks, Collin Burns, Steven Basart, Andy Zou, Mantas Mazeika, Dawn Song, and Jacob Steinhardt.
\newblock Measuring massive multitask language understanding.
\newblock In \emph{9th International Conference on Learning Representations}. OpenReview.net, 2021{\natexlab{a}}.

\bibitem[Hendrycks et~al.(2021{\natexlab{b}})Hendrycks, Burns, Kadavath, Arora, Basart, Tang, Song, and Steinhardt]{DBLP:conf/nips/HendrycksBKABTS21}
Dan Hendrycks, Collin Burns, Saurav Kadavath, Akul Arora, Steven Basart, Eric Tang, Dawn Song, and Jacob Steinhardt.
\newblock Measuring mathematical problem solving with the {MATH} dataset.
\newblock In \emph{Proceedings of the Neural Information Processing Systems Track on Datasets and Benchmarks 1}, 2021{\natexlab{b}}.

\bibitem[Houlsby et~al.(2019)Houlsby, Giurgiu, Jastrzebski, Morrone, de~Laroussilhe, Gesmundo, Attariyan, and Gelly]{DBLP:conf/icml/HoulsbyGJMLGAG19}
Neil Houlsby, Andrei Giurgiu, Stanislaw Jastrzebski, Bruna Morrone, Quentin de~Laroussilhe, Andrea Gesmundo, Mona Attariyan, and Sylvain Gelly.
\newblock Parameter-efficient transfer learning for {NLP}.
\newblock In \emph{Proceedings of the 36th International Conference on Machine Learning}, volume~97 of \emph{Proceedings of Machine Learning Research}, pp.\  2790--2799. {PMLR}, 2019.

\bibitem[Hu et~al.(2022)Hu, Shen, Wallis, Allen{-}Zhu, Li, Wang, Wang, and Chen]{DBLP:conf/iclr/HuSWALWWC22}
Edward~J. Hu, Yelong Shen, Phillip Wallis, Zeyuan Allen{-}Zhu, Yuanzhi Li, Shean Wang, Lu~Wang, and Weizhu Chen.
\newblock Lora: Low-rank adaptation of large language models.
\newblock In \emph{The Tenth International Conference on Learning Representations}. OpenReview.net, 2022.

\bibitem[Ilharco et~al.(2023)Ilharco, Ribeiro, Wortsman, Schmidt, Hajishirzi, and Farhadi]{DBLP:conf/iclr/IlharcoRWSHF23}
Gabriel Ilharco, Marco~T{\'{u}}lio Ribeiro, Mitchell Wortsman, Ludwig Schmidt, Hannaneh Hajishirzi, and Ali Farhadi.
\newblock Editing models with task arithmetic.
\newblock In \emph{The Eleventh International Conference on Learning Representations}. OpenReview.net, 2023.

\bibitem[Jang et~al.(2024)Jang, Yun, and Han]{DBLP:journals/corr/abs-2403-19522}
Dong{-}Hwan Jang, Sangdoo Yun, and Dongyoon Han.
\newblock Model stock: All we need is just a few fine-tuned models.
\newblock \emph{CoRR}, abs/2403.19522, 2024.

\bibitem[Jin et~al.(2023)Jin, Ren, Preotiuc{-}Pietro, and Cheng]{DBLP:conf/iclr/Jin0P023}
Xisen Jin, Xiang Ren, Daniel Preotiuc{-}Pietro, and Pengxiang Cheng.
\newblock Dataless knowledge fusion by merging weights of language models.
\newblock In \emph{The Eleventh International Conference on Learning Representations}. OpenReview.net, 2023.

\bibitem[Ke et~al.(2022)Ke, Lin, Shao, Xu, Shu, and Liu]{DBLP:conf/emnlp/KeLS0SL22}
Zixuan Ke, Haowei Lin, Yijia Shao, Hu~Xu, Lei Shu, and Bing Liu.
\newblock Continual training of language models for few-shot learning.
\newblock In \emph{Proceedings of the 2022 Conference on Empirical Methods in Natural Language Processing}, pp.\  10205--10216. Association for Computational Linguistics, 2022.

\bibitem[Ke et~al.(2023)Ke, Shao, Lin, Konishi, Kim, and Liu]{DBLP:conf/iclr/KeSLKK023}
Zixuan Ke, Yijia Shao, Haowei Lin, Tatsuya Konishi, Gyuhak Kim, and Bing Liu.
\newblock Continual pre-training of language models.
\newblock In \emph{The Eleventh International Conference on Learning Representations}. OpenReview.net, 2023.

\bibitem[Lester et~al.(2021)Lester, Al{-}Rfou, and Constant]{DBLP:conf/emnlp/LesterAC21}
Brian Lester, Rami Al{-}Rfou, and Noah Constant.
\newblock The power of scale for parameter-efficient prompt tuning.
\newblock In \emph{Proceedings of the 2021 Conference on Empirical Methods in Natural Language Processing}, pp.\  3045--3059. Association for Computational Linguistics, 2021.

\bibitem[Li \& Liang(2021)Li and Liang]{DBLP:conf/acl/LiL20}
Xiang~Lisa Li and Percy Liang.
\newblock Prefix-tuning: Optimizing continuous prompts for generation.
\newblock In \emph{Proceedings of the 59th Annual Meeting of the Association for Computational Linguistics and the 11th International Joint Conference on Natural Language Processing}, pp.\  4582--4597. Association for Computational Linguistics, 2021.

\bibitem[Lin et~al.(2022)Lin, Hilton, and Evans]{DBLP:conf/acl/LinHE22}
Stephanie Lin, Jacob Hilton, and Owain Evans.
\newblock Truthfulqa: Measuring how models mimic human falsehoods.
\newblock In \emph{Proceedings of the 60th Annual Meeting of the Association for Computational Linguistics}, pp.\  3214--3252. Association for Computational Linguistics, 2022.

\bibitem[Liu et~al.(2024)Liu, Wang, Yin, Molchanov, Wang, Cheng, and Chen]{DBLP:journals/corr/abs-2402-09353}
Shih{-}Yang Liu, Chien{-}Yi Wang, Hongxu Yin, Pavlo Molchanov, Yu{-}Chiang~Frank Wang, Kwang{-}Ting Cheng, and Min{-}Hung Chen.
\newblock Dora: Weight-decomposed low-rank adaptation.
\newblock \emph{CoRR}, abs/2402.09353, 2024.

\bibitem[Luo et~al.(2023)Luo, Sun, Xu, Zhao, Lou, Tao, Geng, Lin, Chen, and Zhang]{DBLP:journals/corr/abs-2308-09583}
Haipeng Luo, Qingfeng Sun, Can Xu, Pu~Zhao, Jianguang Lou, Chongyang Tao, Xiubo Geng, Qingwei Lin, Shifeng Chen, and Dongmei Zhang.
\newblock Wizardmath: Empowering mathematical reasoning for large language models via reinforced evol-instruct.
\newblock \emph{CoRR}, abs/2308.09583, 2023.

\bibitem[Matena \& Raffel(2022)Matena and Raffel]{DBLP:conf/nips/MatenaR22}
Michael Matena and Colin Raffel.
\newblock Merging models with fisher-weighted averaging.
\newblock In \emph{NeurIPS}, 2022.

\bibitem[Mohammed \& Kora(2023)Mohammed and Kora]{mohammed2023comprehensive}
Ammar Mohammed and Rania Kora.
\newblock A comprehensive review on ensemble deep learning: Opportunities and challenges.
\newblock \emph{Journal of King Saud University-Computer and Information Sciences}, 35\penalty0 (2):\penalty0 757--774, 2023.

\bibitem[Ponti et~al.(2020)Ponti, Glavas, Majewska, Liu, Vulic, and Korhonen]{DBLP:conf/emnlp/PontiGMLVK20}
Edoardo~Maria Ponti, Goran Glavas, Olga Majewska, Qianchu Liu, Ivan Vulic, and Anna Korhonen.
\newblock {XCOPA:} {A} multilingual dataset for causal commonsense reasoning.
\newblock In \emph{Proceedings of the 2020 Conference on Empirical Methods in Natural Language Processing}, pp.\  2362--2376. Association for Computational Linguistics, 2020.

\bibitem[Radford et~al.(2018)Radford, Narasimhan, Salimans, Sutskever, et~al.]{radford2018improving}
Alec Radford, Karthik Narasimhan, Tim Salimans, Ilya Sutskever, et~al.
\newblock Improving language understanding by generative pre-training.
\newblock 2018.

\bibitem[Rafailov et~al.(2023)Rafailov, Sharma, Mitchell, Manning, Ermon, and Finn]{DBLP:conf/nips/RafailovSMMEF23}
Rafael Rafailov, Archit Sharma, Eric Mitchell, Christopher~D. Manning, Stefano Ermon, and Chelsea Finn.
\newblock Direct preference optimization: Your language model is secretly a reward model.
\newblock In \emph{Advances in Neural Information Processing Systems 36}, 2023.

\bibitem[Sakaguchi et~al.(2020)Sakaguchi, Bras, Bhagavatula, and Choi]{DBLP:conf/aaai/SakaguchiBBC20}
Keisuke Sakaguchi, Ronan~Le Bras, Chandra Bhagavatula, and Yejin Choi.
\newblock Winogrande: An adversarial winograd schema challenge at scale.
\newblock In \emph{The Thirty-Fourth {AAAI} Conference on Artificial Intelligence}, pp.\  8732--8740. {AAAI} Press, 2020.

\bibitem[Salimans \& Kingma(2016)Salimans and Kingma]{DBLP:conf/nips/SalimansK16}
Tim Salimans and Diederik~P. Kingma.
\newblock Weight normalization: {A} simple reparameterization to accelerate training of deep neural networks.
\newblock In \emph{Advances in Neural Information Processing Systems 29}, pp.\  901, 2016.

\bibitem[Shoemake(1985)]{DBLP:conf/siggraph/Shoemake85}
Ken Shoemake.
\newblock Animating rotation with quaternion curves.
\newblock In \emph{Proceedings of the 12th Annual Conference on Computer Graphics and Interactive Techniques}, pp.\  245--254. {ACM}, 1985.

\bibitem[Song et~al.(2024)Song, Yu, Li, Yu, Huang, Li, and Wang]{DBLP:conf/aaai/00010LYHLW24}
Feifan Song, Bowen Yu, Minghao Li, Haiyang Yu, Fei Huang, Yongbin Li, and Houfeng Wang.
\newblock Preference ranking optimization for human alignment.
\newblock In \emph{Thirty-Eighth {AAAI} Conference on Artificial Intelligence}, pp.\  18990--18998. {AAAI} Press, 2024.

\bibitem[Touvron et~al.(2023)Touvron, Martin, Stone, Albert, Almahairi, Babaei, Bashlykov, Batra, Bhargava, Bhosale, Bikel, Blecher, Canton{-}Ferrer, Chen, Cucurull, Esiobu, Fernandes, Fu, Fu, Fuller, Gao, Goswami, Goyal, Hartshorn, Hosseini, Hou, Inan, Kardas, Kerkez, Khabsa, Kloumann, Korenev, Koura, Lachaux, Lavril, Lee, Liskovich, Lu, Mao, Martinet, Mihaylov, Mishra, Molybog, Nie, Poulton, Reizenstein, Rungta, Saladi, Schelten, Silva, Smith, Subramanian, Tan, Tang, Taylor, Williams, Kuan, Xu, Yan, Zarov, Zhang, Fan, Kambadur, Narang, Rodriguez, Stojnic, Edunov, and Scialom]{DBLP:journals/corr/abs-2307-09288}
Hugo Touvron, Louis Martin, Kevin Stone, Peter Albert, Amjad Almahairi, Yasmine Babaei, Nikolay Bashlykov, Soumya Batra, Prajjwal Bhargava, Shruti Bhosale, Dan Bikel, Lukas Blecher, Cristian Canton{-}Ferrer, Moya Chen, Guillem Cucurull, David Esiobu, Jude Fernandes, Jeremy Fu, Wenyin Fu, Brian Fuller, Cynthia Gao, Vedanuj Goswami, Naman Goyal, Anthony Hartshorn, Saghar Hosseini, Rui Hou, Hakan Inan, Marcin Kardas, Viktor Kerkez, Madian Khabsa, Isabel Kloumann, Artem Korenev, Punit~Singh Koura, Marie{-}Anne Lachaux, Thibaut Lavril, Jenya Lee, Diana Liskovich, Yinghai Lu, Yuning Mao, Xavier Martinet, Todor Mihaylov, Pushkar Mishra, Igor Molybog, Yixin Nie, Andrew Poulton, Jeremy Reizenstein, Rashi Rungta, Kalyan Saladi, Alan Schelten, Ruan Silva, Eric~Michael Smith, Ranjan Subramanian, Xiaoqing~Ellen Tan, Binh Tang, Ross Taylor, Adina Williams, Jian~Xiang Kuan, Puxin Xu, Zheng Yan, Iliyan Zarov, Yuchen Zhang, Angela Fan, Melanie Kambadur, Sharan Narang, Aur{\'{e}}lien Rodriguez, Robert Stojnic, Sergey Edunov,
  and Thomas Scialom.
\newblock Llama 2: Open foundation and fine-tuned chat models.
\newblock \emph{CoRR}, abs/2307.09288, 2023.

\bibitem[Wang et~al.(2023)Wang, Zhong, Li, Mi, Zeng, Huang, Shang, Jiang, and Liu]{DBLP:journals/corr/abs-2307-12966}
Yufei Wang, Wanjun Zhong, Liangyou Li, Fei Mi, Xingshan Zeng, Wenyong Huang, Lifeng Shang, Xin Jiang, and Qun Liu.
\newblock Aligning large language models with human: {A} survey.
\newblock \emph{CoRR}, abs/2307.12966, 2023.

\bibitem[Wortsman et~al.(2022)Wortsman, Ilharco, Gadre, Roelofs, Lopes, Morcos, Namkoong, Farhadi, Carmon, Kornblith, and Schmidt]{DBLP:conf/icml/WortsmanIGRLMNF22}
Mitchell Wortsman, Gabriel Ilharco, Samir~Yitzhak Gadre, Rebecca Roelofs, Raphael~Gontijo Lopes, Ari~S. Morcos, Hongseok Namkoong, Ali Farhadi, Yair Carmon, Simon Kornblith, and Ludwig Schmidt.
\newblock Model soups: averaging weights of multiple fine-tuned models improves accuracy without increasing inference time.
\newblock In \emph{International Conference on Machine Learning}, volume 162 of \emph{Proceedings of Machine Learning Research}, pp.\  23965--23998. {PMLR}, 2022.

\bibitem[Wu et~al.(2024)Wu, Luo, Li, Pan, Vu, and Haffari]{DBLP:journals/corr/abs-2402-01364}
Tongtong Wu, Linhao Luo, Yuan{-}Fang Li, Shirui Pan, Thuy{-}Trang Vu, and Gholamreza Haffari.
\newblock Continual learning for large language models: {A} survey.
\newblock \emph{CoRR}, abs/2402.01364, 2024.

\bibitem[Xie et~al.(2023)Xie, Aggarwal, and Ahmad]{DBLP:journals/corr/abs-2311-08545}
Yong Xie, Karan Aggarwal, and Aitzaz Ahmad.
\newblock Efficient continual pre-training for building domain specific large language models.
\newblock \emph{CoRR}, abs/2311.08545, 2023.

\bibitem[Xu et~al.(2024)Xu, Sun, Zheng, Geng, Zhao, Feng, Tao, Lin, and Jiang]{xu2024wizardlm}
Can Xu, Qingfeng Sun, Kai Zheng, Xiubo Geng, Pu~Zhao, Jiazhan Feng, Chongyang Tao, Qingwei Lin, and Daxin Jiang.
\newblock Wizard{LM}: Empowering large pre-trained language models to follow complex instructions.
\newblock In \emph{The Twelfth International Conference on Learning Representations}, 2024.

\bibitem[Yadav et~al.(2023)Yadav, Tam, Choshen, Raffel, and Bansal]{DBLP:conf/nips/YadavTCRB23}
Prateek Yadav, Derek Tam, Leshem Choshen, Colin~A. Raffel, and Mohit Bansal.
\newblock Ties-merging: Resolving interference when merging models.
\newblock In \emph{Advances in Neural Information Processing Systems 36}, 2023.

\bibitem[Yu et~al.(2024)Yu, Yu, Yu, Huang, and Li]{yu2023language}
Le~Yu, Bowen Yu, Haiyang Yu, Fei Huang, and Yongbin Li.
\newblock Language models are super mario: Absorbing abilities from homologous models as a free lunch.
\newblock In \emph{International Conference on Machine Learning}. PMLR, 2024.

\bibitem[Yuan et~al.(2023)Yuan, Yuan, Li, Dong, Tan, and Zhou]{DBLP:journals/corr/abs-2308-01825}
Zheng Yuan, Hongyi Yuan, Chengpeng Li, Guanting Dong, Chuanqi Tan, and Chang Zhou.
\newblock Scaling relationship on learning mathematical reasoning with large language models.
\newblock \emph{CoRR}, abs/2308.01825, 2023.

\bibitem[Zellers et~al.(2019)Zellers, Holtzman, Bisk, Farhadi, and Choi]{DBLP:conf/acl/ZellersHBFC19}
Rowan Zellers, Ari Holtzman, Yonatan Bisk, Ali Farhadi, and Yejin Choi.
\newblock Hellaswag: Can a machine really finish your sentence?
\newblock In \emph{Proceedings of the 57th Conference of the Association for Computational Linguistics}, pp.\  4791--4800. Association for Computational Linguistics, 2019.

\bibitem[Zhang et~al.(2023{\natexlab{a}})Zhang, Dong, Li, Zhang, Sun, Wang, Li, Hu, Zhang, Wu, and Wang]{DBLP:journals/corr/abs-2308-10792}
Shengyu Zhang, Linfeng Dong, Xiaoya Li, Sen Zhang, Xiaofei Sun, Shuhe Wang, Jiwei Li, Runyi Hu, Tianwei Zhang, Fei Wu, and Guoyin Wang.
\newblock Instruction tuning for large language models: {A} survey.
\newblock \emph{CoRR}, abs/2308.10792, 2023{\natexlab{a}}.

\bibitem[Zhang et~al.(2023{\natexlab{b}})Zhang, Aljunied, Gao, Chia, and Bing]{DBLP:conf/nips/ZhangAGCB23}
Wenxuan Zhang, Mahani Aljunied, Chang Gao, Yew~Ken Chia, and Lidong Bing.
\newblock M3exam: {A} multilingual, multimodal, multilevel benchmark for examining large language models.
\newblock In \emph{Advances in Neural Information Processing Systems 36}, 2023{\natexlab{b}}.

\bibitem[Zhang \& Yang(2022)Zhang and Yang]{DBLP:journals/tkde/ZhangY22}
Yu~Zhang and Qiang Yang.
\newblock A survey on multi-task learning.
\newblock \emph{{IEEE} Trans. Knowl. Data Eng.}, 34\penalty0 (12):\penalty0 5586--5609, 2022.

\bibitem[Zhao et~al.(2023)Zhao, Zhou, Li, Tang, Wang, Hou, Min, Zhang, Zhang, Dong, Du, Yang, Chen, Chen, Jiang, Ren, Li, Tang, Liu, Liu, Nie, and Wen]{DBLP:journals/corr/abs-2303-18223}
Wayne~Xin Zhao, Kun Zhou, Junyi Li, Tianyi Tang, Xiaolei Wang, Yupeng Hou, Yingqian Min, Beichen Zhang, Junjie Zhang, Zican Dong, Yifan Du, Chen Yang, Yushuo Chen, Zhipeng Chen, Jinhao Jiang, Ruiyang Ren, Yifan Li, Xinyu Tang, Zikang Liu, Peiyu Liu, Jian{-}Yun Nie, and Ji{-}Rong Wen.
\newblock A survey of large language models.
\newblock \emph{CoRR}, abs/2303.18223, 2023.

\end{thebibliography}
\bibliographystyle{iclr2024_conference}

\clearpage
\appendix
\label{section-appendix}

\section{Experimental Details}
\subsection{Details of FT and PT LLMs}\label{section-appendix-llms_backbone_correspondences}
\tabref{tab:llms_backbone_correspondences} depicts the versions and correspondences with backbones of FT and PT LLMs.

\begin{table}[!htbp]
\centering
\caption{Versions and correspondences with backbones of FT and PT LLMs.}
\label{tab:llms_backbone_correspondences}
\begin{tabular}{c|c|c}
\hline
Types                                   & Models                  & Backbones       \\ \hline
FT LLM  & Qwen1.5-7B-Chat\tablefootnote{\url{https://huggingface.co/Qwen/Qwen1.5-7B-Chat}}           & Qwen1.5-7B\tablefootnote{\url{https://huggingface.co/Qwen/Qwen1.5-7B}}\saveFN{\QwenSevenBfn}          \\
                                        PT LLM & Sailor-7B\tablefootnote{\url{https://huggingface.co/sail/Sailor-7B}}         & Qwen1.5-7B\useFN{\QwenSevenBfn} \\ \hline
FT LLM  & Qwen1.5-14B-Chat\tablefootnote{\url{https://huggingface.co/Qwen/Qwen1.5-14B-Chat}}           & Qwen1.5-14B\tablefootnote{\url{https://huggingface.co/Qwen/Qwen1.5-14B}}\saveFN{\QwenFourteenBfn}          \\
                                        PT LLM & Sailor-14B\tablefootnote{\url{https://huggingface.co/sail/Sailor-14B}}         & Qwen1.5-14B\useFN{\QwenFourteenBfn} \\ \hline                           
\multirow{3}{*}{FT LLM}  & WizardLM-13B\tablefootnote{\url{https://huggingface.co/WizardLM/WizardLM-13B-V1.2}}           & Llama-2-13b\tablefootnote{\url{https://huggingface.co/meta-llama/Llama-2-13b-hf}}\saveFN{\llamaTwoThirteenBfn}          \\
                                        & WizardMath-13B\tablefootnote{\url{https://huggingface.co/WizardLM/WizardMath-13B-V1.0}}         & Llama-2-13b\useFN{\llamaTwoThirteenBfn}          \\
                                        & llama-2-13b-code-alpaca\tablefootnote{\url{https://huggingface.co/layoric/llama-2-13b-code-alpaca}}     & Llama-2-13b\useFN{\llamaTwoThirteenBfn}           \\ \hline
\end{tabular}
\end{table}

\subsection{Overview and Evaluation Metrics of Benchmarks}\label{section-appendix-datasets_metrics_details}
The Open LLM Leaderboard is established to assess open-source LLMs using the Eleuther AI Language Model Evaluation Harness \cite{eval-harness}, which encompasses six datasets: AI2 Reasoning Challenge (ARC) \cite{DBLP:journals/corr/abs-1803-05457}, HellaSwag \cite{DBLP:conf/acl/ZellersHBFC19}, MMLU \cite{DBLP:conf/iclr/HendrycksBBZMSS21}, TruthfulQA \cite{DBLP:conf/acl/LinHE22}, Winogrande \cite{DBLP:conf/aaai/SakaguchiBBC20}, and GSM8K \cite{DBLP:journals/corr/abs-2110-14168}. These datasets adopt accuracy as the evaluation metric under various shot settings (25-, 10-, 0-, 5-, 5-, and 5-shot, respectively). The leaderboard ranks models based on the average scores across these six datasets. 

The benchmark for South-East Asian languages is designed with four tasks: XQuAD \cite{DBLP:conf/acl/ArtetxeRY20} (Thai, Vietnamese) and TydiQA \cite{DBLP:journals/tacl/ClarkPNCGCK20} (Indonesian) for question answering; XCOPA \cite{DBLP:conf/emnlp/PontiGMLVK20} (Indonesian, Thai, Vietnamese) for commonsense reasoning; BELEBELE \cite{DBLP:journals/corr/abs-2308-16884} (Indonesian, Thai, and Vietnamese) for reading comprehension; and M3Exam \cite{DBLP:conf/nips/ZhangAGCB23} (Javanese) for examination. All the datasets utilize 3-shot Exact Match (EM) and F1 as evaluation metrics. It is worth noticing that the official code\footnote{\url{https://github.com/sail-sg/sailor-llm}} of Sailor computes multiple metrics for M3Exam on Thai and Vietnamese, which are inconsistent with the originally reported results. Thus, we only present the results of M3Exam (Javanese) in this work.

AlpacaEval 2.0 employs the win rate for assessment, calculated as the proportion of cases where a powerful LLM (GPT-4 Turbo is used in this work) prefers the outputs from the target model over those from GPT-4 Turbo. GSM8K and MATH are evaluated by zero-shot accuracy in addressing mathematical problems. HumanEval and MBPP adopt pass@1 as the evaluation metric, representing the fraction of individually generated code samples that successfully pass the unit tests.

\subsection{Descriptions of Model Merging Baselines}\label{section-appendix-model_merging_methods_descriptions}
We compare with seven commonly-used model merging methods in the experiments:
\begin{itemize}
    \item  \textbf{Average Merging} simply averages the parameters of multiple models for building the merged model \cite{DBLP:conf/icml/WortsmanIGRLMNF22}.
    
    \item \textbf{Task Arithmetic} employs a scaling term to modulate the importance of the backbone and various models to be merged \cite{DBLP:conf/iclr/IlharcoRWSHF23}.

    \item \textbf{SLERP} is tailored for the combination of two models, utilizing spherical interpolation to merge the model weights \cite{DBLP:conf/siggraph/Shoemake85}. 

    \item \textbf{Model Stock} seeks to approximate a center-close weight by considering several FT models, where the backbone is leveraged as an anchor point \cite{DBLP:journals/corr/abs-2403-19522}. 

    \item \textbf{TIES-Merging} aims to mitigate task conflicts in model merging by initially pruning delta parameters with lower magnitudes and subsequently fusing parameters that exhibit consistent signs \cite{DBLP:conf/nips/YadavTCRB23}. 

    \item \textbf{Breadcrumbs} refines model parameters by filtering out the extreme tails (i.e., outliers) in the absolute magnitude distribution of task vectors to derive the final merged model \cite{DBLP:journals/corr/abs-2312-06795}. 

    \item \textbf{DARE} serves as a versatile module for current merging techniques, which first randomly discards delta parameters and then rescales the remaining parameters to preserve the model performance \cite{yu2023language}.
\end{itemize}

\subsection{Analysis of WIDEN and Arithmetic-Based Merging Methods}\label{section-appendix-analysis_widen_arithmetic}
The computation procedure of Average Merging \cite{DBLP:conf/icml/WortsmanIGRLMNF22} for $N$ LLMs is denoted by 
\begin{equation}
   \bm{W}_\text{M} = \frac{1}{N} \sum\limits_{n=1}^N \bm{W}^n = \bm{W}_\text{PRE} + \frac{1}{N} \sum\limits_{n=1}^N \left( \bm{W}^n - \bm{W}_\text{PRE} \right) \in \mathbb{R}^{d \times k}.
\end{equation}
Task Arithmetic \cite{DBLP:conf/iclr/IlharcoRWSHF23} is implemented as follows,
\begin{equation}
   \bm{W}_\text{M} = \bm{W}_\text{PRE} + \lambda \sum\limits_{n=1}^N \left( \bm{W}^n - \bm{W}_\text{PRE} \right) \in \mathbb{R}^{d \times k},
\end{equation}
where $\lambda$ denotes the scaling term. It is straightforward that in \equref{equ:weight_above_average}, if $t$ is set to be minus, all the parameters can be considered crucial, with their importance scores calibrated to $s$. Thus, \equref{equ:weight_adaptive_merging} can be rewritten as
\begin{equation}
   \bm{W}_\text{M} = \bm{W}_\text{PRE} + \sum\limits_{n=1}^N \frac{s + s}{2} \left( \bm{W}^n - \bm{W}_\text{PRE} \right) = \bm{W}_\text{PRE} + s \sum\limits_{n=1}^N \left( \bm{W}^n - \bm{W}_\text{PRE} \right)\in \mathbb{R}^{d \times k}.
\end{equation}
To this end, when $t < 0.0$ and $s = 1/N$, WIDEN transforms into Average Merging; when $t < 0.0$ and $s = \lambda$, WIDEN represents Task Arithmetic.

\subsection{Details of Grid Search on Hyperparameters of Merging Methods}\label{section-appendix-grid_search_model_merging_hyperparameters}
\tabref{tab:hyperparameter_searched_ranges_merging_methods} presents the searched ranges of hyperparameters of model merging approaches.
\begin{table}[!htbp]
\centering
\caption{Hyperparameter searched ranges of model merging approaches.}
\label{tab:hyperparameter_searched_ranges_merging_methods}
\resizebox{1.0\textwidth}{!}
{
\begin{tabular}{c|c}
\hline
Model Merging Methods & Search Ranges of Hyperparameters                                                                                                                                                                                 \\ \hline
Task Arithmetic       & \begin{tabular}[c]{@{}c@{}}scaling term to merge parameters: {[}0.5, 1.0{]}\end{tabular}                                                                                            \\ \hline
SLERP        & \begin{tabular}[c]{@{}c@{}}spherical interpolation factor: {[}0.3, 0.5, 0.7{]}\end{tabular} \\ \hline
Model Stock        & / \\ \hline
TIES-Merging          & \begin{tabular}[c]{@{}c@{}}scaling term to merge parameters: {[}0.5, 1.0{]},\\ ratio to retain parameters with largest-magnitude values: {[}0.5, 0.7, 0.9{]}\end{tabular}         \\ \hline
Breadcrumbs               & \begin{tabular}[c]{@{}c@{}}scaling term to merge parameters: {[}0.5, 1.0{]},\\ ratio to mask parameters with largest-magnitude values: {[}0.01, 0.05{]},\\ ratio to retain parameters {[}0.9{]}\end{tabular}  \\ \hline
WIDEN          & \begin{tabular}[c]{@{}c@{}}factor to indicate the multiple above the average: {[}1.0, 2.0{]},\\ factor to calibrate scores: {[}1.0{]}\end{tabular}         \\ \hline
\end{tabular}
}
\end{table}

\subsection{Parameter Changed Ranges of FT and PT LLMs}\label{section-appendix-llms_delta_parameter_changed_ranges}
We depict the statistics about the deciles of parameter changed ranges of both FT and PT LLMs in \tabref{tab:statistics_parameters_changed_ranges}, which are derived by first sorting the entire ranges and then indexing at positions corresponding to 0\%, 10\%, 20\%, ..., 100\%.

\begin{table}[!htbp]
\centering
\caption{Statistics about the deciles of parameter changed ranges of FT and PT LLMs.}
\label{tab:statistics_parameters_changed_ranges}
\resizebox{1.0\textwidth}{!}
{
\setlength{\tabcolsep}{0.6mm}
{
\begin{tabular}{cc|ccccccccccc}
\hline
\multicolumn{2}{c|}{Models}                                                                                            & 0\% (min)         & 10\%        & 20\%        & 30\%        & 40\%        & 50\%       & 60\%       & 70\%       & 80\%       & 90\%       & 100\% (max)        \\ \hline
\multicolumn{2}{c|}{\begin{tabular}[c]{@{}c@{}}Qwen1.5-7B-Chat\\ vs. Qwen1.5-7B\end{tabular}} & -2.43e-02 & -4.27e-04 & -2.44e-04 & -1.22e-04 & -3.05e-05 & 0.00 &  3.05e-05 &  1.22e-04 &  2.44e-04 &  4.27e-04 & 2.29e-02 \\ \hline
\multicolumn{2}{c|}{\begin{tabular}[c]{@{}c@{}}Sailor-7B\\ vs. Qwen1.5-7B\end{tabular}} & -0.27 & -0.57e-02 & -0.37e-02 & -0.23e-02 & -0.11e-02 &  0.00 &  0.11e-02 &  0.23e-02 & 0.37e-02 &  0.57e-02 &  0.25     \\ \hline
\multicolumn{2}{c|}{\begin{tabular}[c]{@{}c@{}}Qwen1.5-14B-Chat\\ vs. Qwen1.5-14B\end{tabular}} & -2.34e-02 & -4.27e-04 & -2.44e-04 & -1.22e-04 & -3.05e-05 & 0.00 &  3.05e-05 &  1.22e-04 &  2.44e-04 &  4.27e-04 & 2.06e-02 \\ \hline
\multicolumn{2}{c|}{\begin{tabular}[c]{@{}c@{}}Sailor-14B\\ vs. Qwen1.5-14B\end{tabular}} & -0.36 & -0.78e-02 & -0.51e-02 & -0.31e-02 & -0.15e-02 &  0.00 &  0.15e-02 &  0.31e-02 & 0.51e-02 &  0.78e-02 &  0.42     \\ \hline
\multicolumn{2}{c|}{\begin{tabular}[c]{@{}c@{}}WizardLM-13B\\ vs. Llama-2-13b\end{tabular}}                    & -3.93e-02 & -0.16e-02 & -0.10e-02 & -0.06e-02 & -0.03e-02 &  0.00 &  0.03e-02 &  0.06e-02 & 0.10e-02 &  0.16e-02 &  4.81e-02     \\ \hline
\multicolumn{2}{c|}{\begin{tabular}[c]{@{}c@{}}WizardMath-13B\\ vs. Llama-2-13b\end{tabular}}                  & -0.69e-02 & -0.06e-02 & -0.04e-02 & -0.02e-02 & -0.01e-02 &  0.00 &  0.01e-02 &  0.02e-02 & 0.04e-02 &  0.06e-02 &  0.74e-02     \\ \hline
\multicolumn{2}{c|}{\begin{tabular}[c]{@{}c@{}}llama-2-13b-code-alpaca\\ vs. Llama-2-13b\end{tabular}}              & -8.42e-02 & -3.05e-05 &  0.00 &  0.00 &  0.00 & 0.00 &  0.00 &  0.00 &  0.00 &  3.05e-05 & 7.98e-02 \\ \hline
\end{tabular}
}
}
\end{table}

\end{document}